\documentclass[journal]{IEEEtran}
\usepackage{amsmath,amsfonts}
\usepackage{algorithmic}
\usepackage{algorithm}
\usepackage{array}
\usepackage{subcaption}
\usepackage{textcomp}
\usepackage{stfloats}
\usepackage{url}
\usepackage{verbatim}
\usepackage{graphicx}

\usepackage{multirow}    
\usepackage{graphicx}    
\usepackage{array}      
\usepackage{amssymb}    
\usepackage[numbers]{natbib}
\usepackage{booktabs}
\usepackage{threeparttable}
\usepackage{makecell}
\begin{document}

\title{Learning Language-Driven Sequence-Level Modal-Invariant Representations for Video-Based Visible-Infrared Person Re-Identification}

\author{Xiaomei~Yang,~\IEEEmembership{}
        Antai~Liu,~\IEEEmembership{}
        Xizhan~Gao,~\IEEEmembership{}
        Fa~Zhu,~\IEEEmembership{}
        Sijie~Niu,~\IEEEmembership{Member,~IEEE,}
        and~Giancarlo~Fortino,~\IEEEmembership{Fellow,~IEEE}
\thanks{Xiaomei Yang, Antai Liu, Xizhan Gao, Sijie Niu are with the Shandong Key Laboratory of Ubiquitous Intelligent Computing, School of Information Science and Engineering, University of
Jinan, Jinan 250022, China (e-mail: ise\_gaoxz@ujn.edu.cn). \emph{(Corresponding author: Xizhan Gao.)}}
\thanks{Fa Zhu is with the College of Information Science and Technology \& Artificial Intelligence, Nanjing Forestry University, Nanjing 210037, China.}
\thanks{Giancarlo Fortino is with the Department of Informatics, Modeling, Electronics, and Systems, University of Calabria, Via P. Bucci, cubo 41C, 87036, Rende (CS), Italy.}
}

\markboth{Journal of \LaTeX\ Class Files,~Vol.~~, No.~~, June~2026}%
{Shell \MakeLowercase{\textit{et al.}}: A Sample Article Using IEEEtran.cls for IEEE Journals}


\maketitle

\begin{abstract}
The core of video-based visible-infrared person re-identification (VVI-ReID) lies in learning sequence-level modal-invariant representations across different modalities. Recent research tends to use modality-shared language prompts generated by CLIP model to guide the learning of modal-invariant representations. Despite achieving optimal performance, such methods still face limitations in efficient spatial-temporal modeling, sufficient cross-modal interaction, and explicit modality-level loss guidance. To address these issues, we propose a novel VVI-ReID method, language-driven sequence-level modal-invariant representation learning (LSMRL), which consists of three modules: the spatial-temporal feature learning (STFL) module, the semantic diffusion (SD) module and the cross-modal interaction (CMI) module. More specifically, to enable parameter- and computationally efficient spatial-temporal modeling, we design the STFL module, which is built upon CLIP with minimal modifications. This module decomposes CLIP's vision encoder into a basic vision encoder, a spatial-temporal grouped encoder, and a temporal patch shift encoder, which jointly capture fine-grained spatial structures and dynamic temporal dependencies without excessive computational overhead. To achieve sufficient cross-modal interaction and enhance the learning of modal-invariant features, the SD module is proposed to diffuse modality-shared language prompts into visible and infrared features to establish preliminary modal consistency. Subsequently, the CMI module is further developed to leverage bidirectional cross-modal self-attention to eliminate residual modality gaps and refine sequence-level modal-invariant representations. To explicitly enhance the learning of modal-invariant representations, we construct a multi-loss system that jointly utilizes identity- and modality-level losses to improve the features' discriminative ability and their generalization to unseen categories. Extensive experiments on large-scale VVI-ReID datasets demonstrate the superiority of LSMRL over state-of-the-art methods. The code will be available at https://github.com/y0406/LSMRL.
\end{abstract}
\begin{IEEEkeywords}
Video-based visible-infrared person Re-ID, Cross-modality interaction, Visual-language model CLIP, Modal-invariant representation, Efficient spatial-temporal modeling
\end{IEEEkeywords}
\section{Introduction}
\IEEEPARstart{V}{ideo-based} person re-identification (V-ReID) is one of the key technologies in the field of intelligent visual surveillance, whose core goal is to accurately match the identity information of the same pedestrian across video sequences from different cameras and scenarios \citep{Li2026LSTARS, Wang2025TAE}. Compared with traditional image-based person re-identification (ReID) task \citep{Lin2026VLM, Liu2026MDEL}, V-ReID can fully leverage spatial-temporal dynamic information in video sequences, effectively compensating for the feature loss issues in single images caused by occlusion, sudden changes in illumination, and view angle shifts. With the breakthroughs in deep learning technology and the establishment of large-scale annotated datasets \citep{zheng2016mars}, research on V-ReID has achieved significant progress. Despite this progress, existing V-ReID methods \citep{MA2024128479} rely heavily on visible light and fail in low-light/harsh weather. Fortunately, infrared cameras, which image by capturing thermal radiation, can effectively solve this issue, thus giving rise to the task of video-based visible-infrared person re-identification (VVI-ReID). VVI-ReID enables pedestrian matching between visible and infrared videos across varying lighting conditions, thereby supporting 24/7 monitoring requirements.

\begin{figure}[!t]
	\centering
	\includegraphics[width=0.5\textwidth]{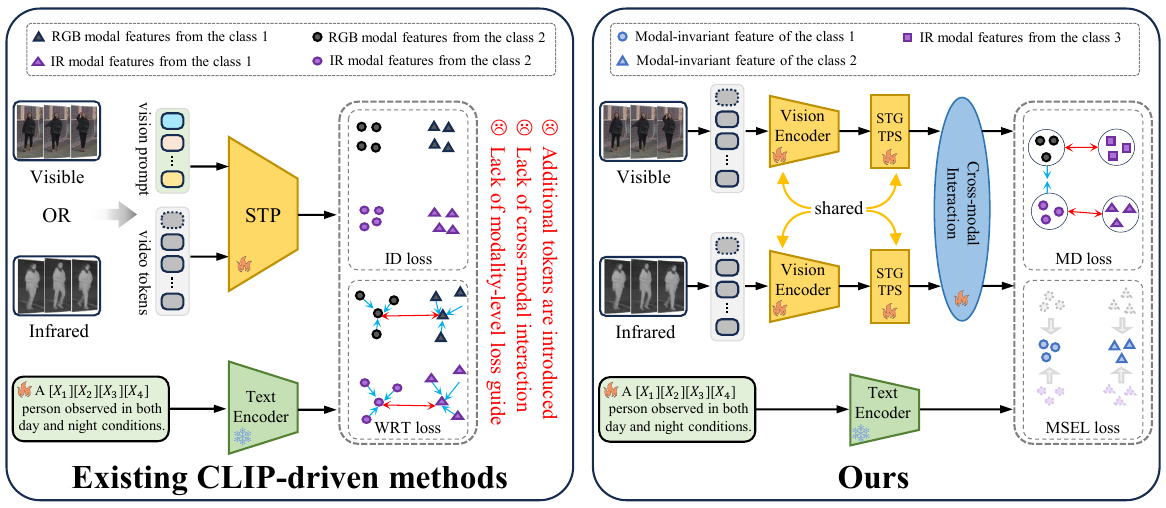}
	\caption{The motivation of this paper: (a) Existing method introduces additional computational overhead when modeling spatial-temporal features, and it struggles to learn robust modal-invariant representations due to the lack of cross-modal interaction and modality-level loss guidance. (b) Our method designs a low-cost STFL module to efficiently model spatial-temporal information, constructs SD and CMI modules to achieve sufficient cross-modal interaction, and utilizes MD and MSEL losses to explicitly enhance cross-modal consistency. }
	\label{fig:motivation}
\vspace{-8pt}
\end{figure} 

Unlike single modality V-ReID, VVI-ReID aims to match pedestrian identities across visible (RGB) and infrared (IR) video sequences. However, VVI-ReID faces two core challenges: the inherent modality gap between visible and infrared data 
and the need to effectively model spatial-temporal dependencies in cross-modal video sequences. 
To address these challenges, a series of VVI-ReID methods \citep{lin2022learning, zhou2023video, du2023video, Wang2025DIRL, Li2023IBAN, feng2024cross} have emerged, which can be categorized into two types: CNN-based methods and Transformer-based methods. CNN-based methods typically adopt a shared CNN network to learn spatial features of video frames, and utilize average pooling \citep{lin2022learning, zhou2023video, du2023video, Wang2025DIRL} or LSTM \citep{Li2023IBAN} for spatial-temporal feature aggregation. This type of method excels at capturing local spatial features but suffers from limitations in modeling long-range temporal dependencies. Transformer-based methods \citep{feng2024cross} first tokenize a video clip into multiple 3D tubes, and then use a Transformer network to model local spatial-temporal features and long-range temporal dependencies. Despite the certain effectiveness of these methods, ReID is inherently a fine-grained task, existing VVI-ReID methods adopt coarse-grained one-hot labels as supervisory signals, failing to provide sufficient fine-grained guidance \citep{li2025video}. Furthermore, these methods rely solely on video data for model training, resulting in extracted visual features lacking high-level semantic information \citep{li2023clip}.

Recently, the vision-language learning paradigm has garnered significant attention due to its robust capacity for extracting semantically rich visual features. As a representative vision-language pre-training model \citep{radford2021learning}, CLIP has attained remarkable performance across a wide range of downstream computer vision tasks \citep{Wu2024BOV} and has also been successfully applied to the person ReID field, including image-based ReID methods \citep{li2023clip, Yu2024CSDN, yang2025clip4vireid, Hu2025CLIPMC,Chen2023CCLNet} and video-based ReID methods  \citep{yu2024tf, li2025video, yu2025xreid}. The former is represented by CLIP-ReID \citep{li2023clip}, CSDN \citep{Yu2024CSDN}, CLIP-MC \citep{Hu2025CLIPMC}, CLIP4VI-ReID \citep{yang2025clip4vireid}, et al., which transfer the CLIP model to image-based ReID tasks through diverse technical routes. These studies demonstrate that the synergy between visual information and language descriptions enables the model to acquire high-level semantic representations associated with target pedestrians. However, these methods are designed specifically for image data and thus cannot effectively handle video-based ReID tasks. For the latter (video-based ReID methods), representative approaches include TF-CLIP \citep{yu2024tf} and VLD \citep{li2025video}, which extend the CLIP framework to temporal modeling. Despite yielding promising improvements, TF-CLIP is a single model V-ReID method and thus cannot address the cross-modal VVI-ReID task. Although VLD is specifically designed for VVI-ReID, it still suffers from the following limitations (as illustrated in Fig. \ref{fig:motivation}): (1) It facilitates spatial-temporal information exchange via spatial-temporal hubs, which requires introducing additional tokens into CLIP's vision encoder. This inevitably increases computational overhead, as the computational complexity of the encoder layers scales quadratically with the number of input tokens. (2) A shared network is employed to learn modal-invariant features, but the critical role of cross-modal feature interaction is overlooked, which renders the acquired modal-invariant features insufficiently robust. (3) Exclusively relying on identity-level loss to enhance feature discriminability, it fails to incorporate modality-level loss for explicitly reinforcing cross-modal consistency. This oversight results in the learned features suffering from inadequate modal alignment, thereby hindering the model's performance on cross-modal matching tasks.

To address the aforementioned issues, this paper proposes the language-driven sequence-level modal-invariant representation learning method (LSMRL) for VVI-ReID. As illustrated in Fig. \ref{fig:frame}, LSMRL is structured into three mutually complementary modules: the spatial-temporal feature learning (STFL) module, the semantic diffusion (SD) module, and the cross-modal interaction (CMI) module. Firstly, the STFL module is used to efficiently model the spatial-temporal features of video sequences from both modalities. Built upon the CLIP with minimal modifications, this module not only accurately captures the spatial-temporal dynamic information but also effectively avoids additional computational overhead. Then, the SD and CMI modules are jointly used to achieve coarse-to-fine cross-modal interaction. The SD module employs a semantic diffusion mechanism to inject modality-shared text semantics into RGB and IR modality features, thereby realizing coarse-grained cross-modal interaction. Building on this foundation, the CMI module leverages a self-attention mechanism to enable bidirectional cross-modal interaction, further mitigating modality discrepancies and obtaining sequence-level modal-invariant features. Finally, modality-level loss functions such as MD and MSEL are used to explicitly guide the learning of modal-invariant features, further enhancing the features' discriminative ability and their generalization to unseen categories. The main contributions of this paper are summarized as follows:
\begin{enumerate}
  \item We propose a novel VVI-ReID method namely LSMRL, which leverages CLIP's cross-modal semantic alignment capability to guide the learning of sequence-level modal-invariant features, while addressing existing limitations in efficient spatial-temporal modeling, sufficient cross-modal interaction, and explicit modality-level loss guidance.
  \item We design the STFL module with a parameter- and computationally efficient structure. By reusing CLIP's pre-trained Transformer blocks and introducing spatial-temporal grouped attention and the temporal patch shift mechanism, it achieves high-performance spatial-temporal modeling with limited extra computational overhead.
  \item We design a two-stage cross-modal interaction mechanism with SD and CMI modules. SD bridges semantic gaps via text semantics, CMI refines modal consistency through bidirectional interaction, and this dual mechanism reduces RGB/IR-induced modality gaps while improving pedestrian identity feature discriminability.
  \item Extensive experiments on large-scale VVI-ReID datasets demonstrate that LSMRL outperforms state-of-the-art (SOTA) methods. For instance, on the BUPTCampus dataset, it achieves 71.1\% Rank-1 and 67.5\% mAP in the infrared-to-visible task, surpassing the SOTA method by 5.8 and 4.0 percentage points.
\end{enumerate}

The rest of this paper is arranged as follows. We review the related works in Sec. \ref{sec:II}, and provide the details of LSMRL in Sec. \ref{sec:III}. In Sec. \ref{sec:IV}, we report the experimental results of LSMRL on the VVI-ReID task. Finally, we conclude this work and discuss the potential improvements in Sec. \ref{sec:V}.


\section{Relate Work} \label{sec:II}
\subsection{Video-based Visible-Infrared Person Re-Identification}
To address the challenges of cross-modal feature alignment and spatial-temporal feature modeling faced by VVI-ReID, numerous VVI-ReID methods have emerged, which can be categorized into two types: CNN-based methods \citep{lin2022learning, zhou2023video, du2023video, Wang2025DIRL, Li2023IBAN} and Transformer-based methods \citep{feng2024cross}. For instance, Lin et al. \citep{lin2022learning} proposed the first VVI-ReID method, MITML, which uses CNN and LSTM to learn spatial and temporal features respectively and reduces the modality gap by learning modal-invariant features. Similar to MITML, IBAN \citep{Li2023IBAN} also relies on CNN and LSTM to model spatial-temporal features, and leverages anaglyph as an intermediary to learn modality-irrelevant features. Also based on the CNN architecture, SAADG \citep{zhou2023video} focuses on eliminating distractors within the modality and further exploring the correlations among videos of the same category. Du et al. \citep{du2023video} proposed the second VVI-ReID dataset and presented the AuxNet method, which uses CNN and temporal average pooling to obtain spatial-temporal features and leverages GAN networks to mitigate the modality gap. The DIRL method proposed by Wang et al. \citep{Wang2025DIRL}  first uses a shared CNN network to capture frame-level modal-invariant features, then adopts a cross-modal attention mechanism to achieve cross-modal interaction and the learning of sequence-level modal-invariant features, and finally removes residual modality-related information from the modal-invariant features via a decoupling module. Feng et al. \citep{feng2024cross} introduced the Transformer architecture into the VVI-ReID field and proposed the CST method, which leverages cross-frame and multi-frame Transformer modules to achieve spatial-temporal feature modeling and modal-invariant feature learning. Although certain progress has been made, existing methods only rely on video data and one-hot labels to perform VVI-ReID tasks, while overlooking the potential of language or text descriptions in terms of fine-grained, sequence-level, and cross-modal representations.

\subsection{Person ReID with CLIP}
Benefiting from its robust capability to extract semantically rich visual features, the vision-language model CLIP has been applied to the field of person re-identification, and a series of CLIP-based ReID methods have emerged, including image-based ReID methods \citep{li2023clip, Yu2024CSDN, yang2025clip4vireid, Hu2025CLIPMC, Chen2023CCLNet, Li2025UDG, Wang2025SVLLReID} and video-based ReID methods  \citep{yu2024tf, Yu2025CLIMB-ReID, li2025video}. Li et al. \citep{li2023clip} pioneered the application of CLIP to the ReID field and proposed the CLIP-ReID method, which demonstrates that the synergy between visual information and language descriptions can effectively boost image ReID performance. Wang et al. \citep{Wang2025SVLLReID} proposed the SVLL-ReID method, which explores whether self-supervision can facilitate the application of CLIP to image ReID tasks. The U-DG method proposed by Li et al. \citep{Li2025UDG} leverages CLIP to address the ReID task with cross-camera unpaired samples. Subsequently, CLIP was extended to the image-based visible-infrared ReID field. CSDN \citep{Yu2024CSDN} learns language prompts for RGB and IR images respectively, and aligns RGB-IR modality features by leveraging the fused text information. In contrast, Yang et al. \citep{yang2025clip4vireid} argued that learning language prompts for IR-modal images would introduce noise interference, and thus proposed the CLIP4VI-ReID method, which only learns prompts for RGB images and gradually achieves feature alignment through a three-stage training process. Hu et al. \citep{Hu2025CLIPMC} and Chen et al. \citep{Chen2023CCLNet}, proceeding from the perspectives of modality compensation and unsupervised learning respectively, proposed the CLIP-MC and USL-VI-ReID methods, thereby further expanding the application scope of CLIP. To apply CLIP to video-based ReID tasks, Yu et al. successively proposed two text-free methods, namely TF-CLIP \citep{yu2024tf} and CLIMB-ReID \citep{Yu2025CLIMB-ReID}, demonstrating that CLIP delivers excellent performance in temporal modeling tasks. The VLD method proposed by Li et al. \citep{li2025video} further extends CLIP to the VVI-ReID field. It captures spatial-temporal features via a spatial-temporal hub and learns modal-invariant features through a shared network architecture. Despite the remarkable success achieved by the aforementioned methods, the potential of CLIP in VVI-ReID tasks remains to be fully explored, and existing methods still have room for improvement in terms of efficient spatial-temporal modeling and modal-invariant feature learning. In view of this, we propose a novel CLIP-driven modal-invariant representation learning network for VVI-ReID tasks.
 
 
\begin{figure*}[!t]
	\centering
	\includegraphics[width=0.9\textwidth]{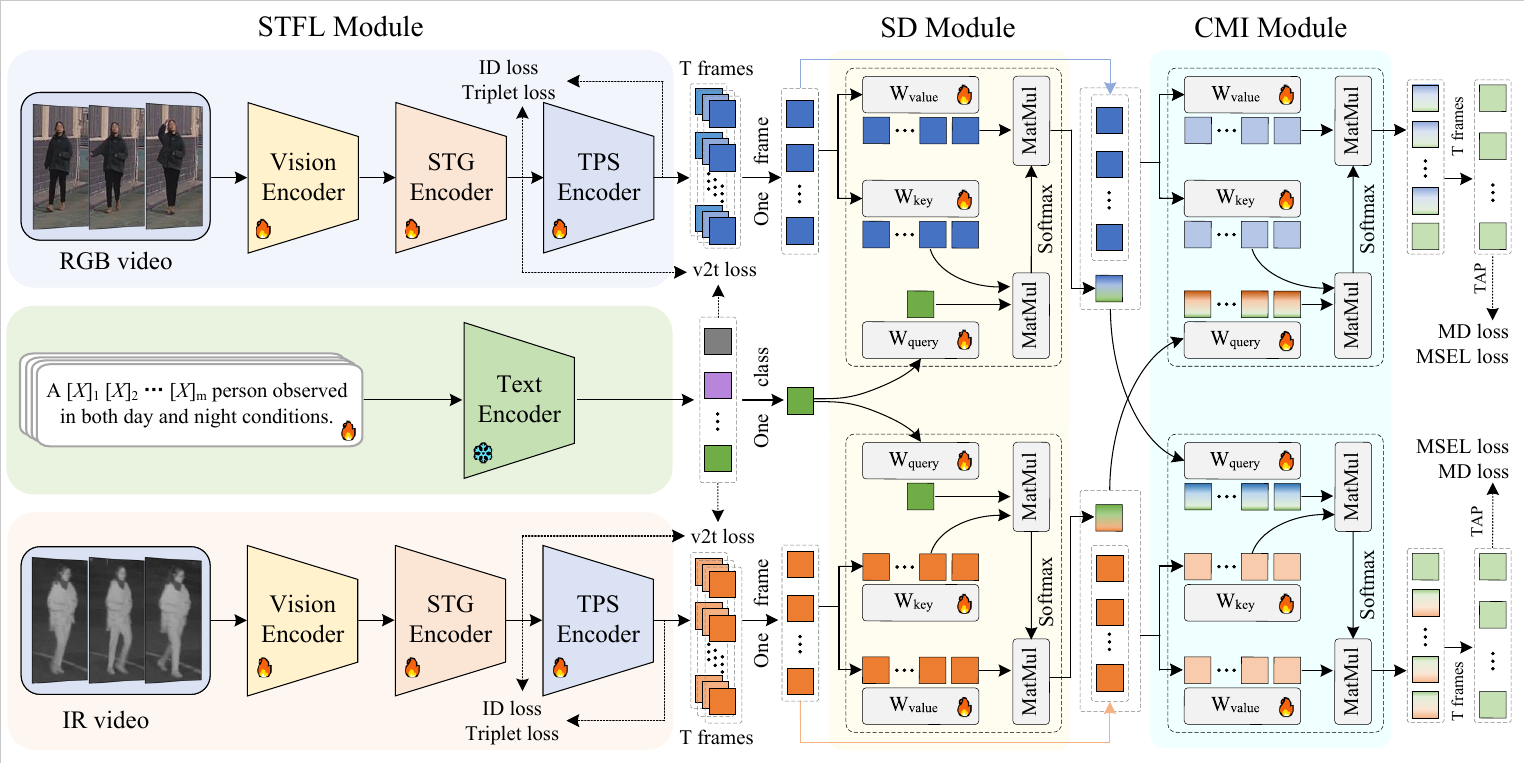}
	\caption{The architecture of the LSMRL method, which consists of the STFL, SD, and CMI modules. The STFL module is first utilized to efficiently model spatial-temporal information of pedestrian sequences with limited extra computational overhead. Then the SD module is applied to diffuse modality-shared text semantics into RGB and IR features, laying the foundation for modal-invariant feature learning. The CMI module is finally used to enable bidirectional cross-modal feature interaction, further eliminating the modality gap and refining discriminative sequence-level modal-invariant representations.  }
	\label{fig:frame}
\end{figure*} 
\section{Methodology} \label{sec:III}
The objective of the VVI-ReID task is to learn sequence-level modal-invariant pedestrian representations that are not only discriminative for identity but also consistent across modalities, thereby enabling accurate cross-modal video matching. To achieve this goal, we propose the LSMRL approach. As illustrated in Fig. \ref{fig:frame}, LSMRL consists of three modules: the spatial-temporal feature learning module, the semantic diffusion module and the cross-modal interaction module. First, to address the challenge of handling diverse data from different modalities, we design the STFL module based on the CLIP model. By integrating spatial-temporal grouped attention and the temporal patch shift mechanism, this module efficiently models spatial-temporal relationships and obtains sequence-level features. Then, we utilize the SD module to diffuse the modality-shared text semantics to the RGB and IR modality features, thereby obtaining preliminary modal-invariant features and laying the groundwork for more robust cross-modal interactions in subsequent modules. Finally, we use the CMI module to achieve RGB-IR cross-modal interaction, further eliminating the modality gap in CLIP's multi-modal space and obtaining the final sequence-level modal-invariant features with stronger discriminative ability. Next, we will elaborate on each module in detail.

\subsection{Spatial-Temporal Feature Learning Module}
In the task of VVI-ReID, the input consists of two modal video sequences, and each sequence is composed of multiple frames that capture the temporal dynamics of the same pedestrian. More specifically, the input contains the RGB sequence $R = \{R_t \mid R_t \in \mathbb{R}^{H \times W \times C} \}_{t=1}^T $ and the IR sequence $I = \{I_t \mid I_t \in \mathbb{R}^{H \times W \times C} \}_{t=1}^T$,
where $R_t$ and $I_t$ denote the $t$-th frame of the RGB and IR sequences, respectively, $T$ denotes the number of frames, and $H,W,C$ represent the height, width, and channel of each frame, respectively. 

As the foundation of sequence-level modal-invariant feature learning, the STFL module is designed to address two key challenges: (1) adapting CLIP's image-centric pre-trained encoder to video data in a parameter- and computationally efficient manner; (2) comprehensively capturing both spatial structure information and temporal dynamic characteristics of pedestrians. To this end, we decompose and extend CLIP's vision encoder into three functionally complementary sub-modules: the basic vision encoder, the spatial-temporal grouped (STG) encoder, and the temporal patch shift (TPS) encoder, which work sequentially to generate high-quality sequence-level features without excessive computational overhead.


\textbf{Vision Encoder} \ \ We construct the basic vision encoder using the first $8$ Transformer blocks of the original CLIP vision encoder. The core rationale is that CLIP's pre-training on massive image-text pairs endows these early-to-middle blocks with the ability to capture generic, fine-grained visual semantics (e.g., pedestrian contours, clothing textures, and body part details) that are widely applicable across RGB and IR modalities. By directly reusing these pre-trained blocks, we avoid redundant parameter training while preserving valuable semantic knowledge—ensuring each frame in the video sequence is converted into a high-quality frame-level feature with basic identity discriminability. This sub-module serves as the ``feature foundation'' of the STFL module, as reliable frame-level representations are prerequisite for subsequent spatial-temporal fusion. Let $f_1(\cdot)$ denotes the basic vision encoder, then it outputs $\bar{R}_t = f_1(R_t^{\prime})$, where $R_t^{\prime} = [cls, {r_t^1}^{\prime}, {r_t^2}^{\prime}, \cdots, {r_t^N}^{\prime}]$ is the feature after patch division and position embedding, $\bar{R}_t = [cls, \bar{r}_t^1, \bar{r}_t^2, \cdots, \bar{r}_t^N]$, and $N$ is the number of patches. Similarly, we can obtain $\bar{I}_t = f_1(I_t^{\prime})$.

\begin{figure}[!t]
\centering
\includegraphics[width=3.5 in]{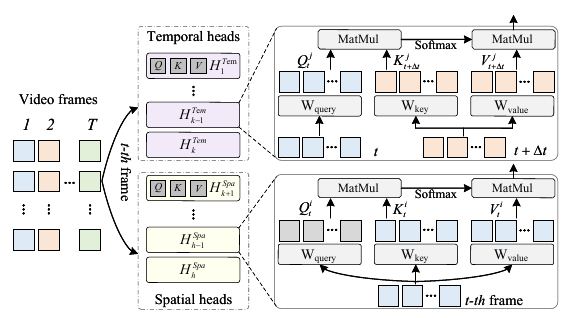}
\caption{Diagram of the STG encoder. In MHSA, the $h$ heads are split into two groups: $k$ temporal heads and $h-k$ spatial heads. Here, $H^{Tem}$ denotes temporal head, $H^{Spa}$ denotes spatial head.}
\label{STG}
\vspace{-8pt}
\end{figure}

\textbf{STG Encoder} \ \ Transferring image models (e.g., CLIP) to video tasks typically requires the incorporation of additional temporal modeling modules, which not only introduces extra parameters and computational overhead but also incurs higher training costs. To address this issue while fully exploiting the sequential nature of video data (which is underemphasized in CLIP's original design), inspired by literatures \citep{li2024zeroi2v, Wu2024OpenVCLIP}, we adopt the last $4$ Transformer blocks of CLIP as the STG encoder and modify their multi-head self-attention (MHSA) mechanism via dividing all heads into two groups. Specifically, as shown in Fig. \ref{STG}, assuming there are $h$ heads in each block, we use the first $k$ heads as the temporal heads and treat the remaining $h-k$ heads as spatial heads.

\emph{Temporal Heads}: Suppose $1 \leq j \leq k$, the $j$-th temporal head can be computed as follows. Considering the RGB modality input $\bar{R}_t$, we can calculate the query matrix at time $t$ by
\begin{equation}
\begin{aligned}
Q_t^j = W_{t,q}^j \cdot \bar{R}_t,
\end{aligned}
\label{eq:1}
\end{equation}
where $W_{t,q}^j$ is a learnable parameter, $Q_t^j$ represents the query matrix corresponding to the $j$-th head on $t$-frame. We can also calculate the key and value matrices at $t+\Delta t_j$ time by 
\begin{equation}
\begin{aligned}
& K_{t+\Delta t_j}^j = W_{t+\Delta t_j,k}^j \cdot \bar{R}_{t+\Delta t_j}, \\
& V_{t+\Delta t_j}^j = W_{t+\Delta t_j,v}^j \cdot \bar{R}_{t+\Delta t_j},
\end{aligned}
\label{eq:2}
\end{equation}
where $\Delta t_j$ represents the time offset, which is used to achieve cross-time interaction. Then, the $j$-th temporal head outputs
\begin{equation}
\begin{aligned}
T_{head}^j = Att(Q_t^j, K_{t+\Delta t_j}^j, V_{t+\Delta t_j}^j), \ j=1, \cdots, k.
\end{aligned}
\label{eq:3}
\end{equation}
At this point, each image patch captures information not only from its own frame but also from a neighboring frame \citep{Wu2024OpenVCLIP}. The temporal heads redefine the attention scope as inter-frame interaction, enabling explicit modeling of the temporal dynamics unique to video sequences (such as posture changes and subtle motion patterns) to effectively distinguish pedestrians with similar appearances. Furthermore, this design does not introduce additional parameters or tokens during the computation of each attention head, thereby ensuring that the model's parameter count and computational complexity remain unchanged.

\emph{Spatial Heads}: Suppose ${k+1} \leq i \leq h$, the $i$-th spatial head first calculates the $Q_t^i, K_t^i, V_t^i$ matrices at time $t$, then it outputs
\begin{equation}
\begin{aligned}
S_{head}^i = Att(Q_t^i, K_t^i, V_t^i), \ i=k+1, \cdots, h.
\end{aligned}
\label{eq:4}
\end{equation}
The spatial heads retain the original intra-frame attention scope and focus on refining spatial relationships (e.g., the relative positions of a pedestrian's head, torso, and limbs). By enhancing the representation of identity-related spatial regions and suppressing background clutter, they play a crucial role in mitigating the impact of scene variations in cross-modal scenarios.

Finally, the output of STG encoder can be expressed as
\begin{equation}
\begin{aligned}
\tilde{R}_t = Concat(T_{head}^1, \cdots, T_{head}^k, S_{head}^{k+1}, \cdots, S_{head}^{h}) \cdot W_{att}^O,
\end{aligned}
\label{eq:5}
\end{equation}
where $W_{att}^O$ denotes the dimensionality reduction matrix, which projects the token dimension back to ${N+1}$. In the same manner, for the IR modality input $\bar{I}_t$, the STG encoder outputs $\tilde{I}_t = f_2(\bar{I}_t)$, where $f_2(\cdot)$ denotes the STG encoder. By separating spatial and temporal modeling into dedicated groups, the STG encoder avoids mutual interference between the two types of dependencies (a limitation of vanilla full-attention) while reusing CLIP’s pre-trained block parameters—achieving targeted spatial-temporal learning without increasing computational complexity.

\begin{figure}[!t]
\centering
\includegraphics[width=3.5 in]{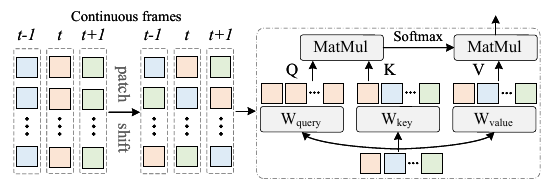}
\caption{Diagram of the TPS encoder. Tokens from neighboring frames are shifted along the temporal dimension, and spatial-temporal modeling is achieved through the spatial-only Transformer layer, with the temporal length being manually controllable.}
\label{TPS}
\vspace{-8pt}
\end{figure}

\textbf{TPS Encoder} \ \ Although the STG encoder has a certain ability to learn temporal features, each of its temporal heads can only enable interaction between two frames, making it difficult to capture long-range cross-frame dependencies. Meanwhile, it lacks flexibility in modeling sparse temporal interactions. Therefore, to efficiently model long-range temporal correlations and enhance sparse spatial-temporal interactions, we add an additional transformer block (with the same parameter configuration as CLIP's blocks for efficiency) as the TPS encoder, which integrates the temporal patch shift mechanism \citep{xiang2022TPS}. 

More specifically, for the sequence of features $\{ \tilde{R}_t \}_{t=1}^T$ output by the STG encoder, we perform cyclic shifting on frame tokens along the temporal dimension using the following formula, which explicitly enhances the correlation between neighboring frames while keeping the network structure unchanged:
\begin{equation}
\begin{aligned}
Z_t^R = A \odot \tilde{R}_{t-1} + B \odot \tilde{R}_t + C \odot \tilde{R}_{t+1},
\end{aligned}
\label{eq:6}
\end{equation}
where $A = [a_0, a_1, \cdots, a_N]$, $B = [b_0, b_1, \cdots, b_N]$, $C = [c_0, c_1, \cdots, c_N]$ are token shift matrices, $a_i, b_i, c_i$ denote vectors of all ones or all zeros, and they satisfy $a_i+b_i+c_i =1$. Here, the length of neighboring frames can be manually controlled, and longer neighboring frames can model long-range cross-frame dependencies.

As shown in Fig. \ref{TPS}, after obtaining the shifted feature,  we use $Z_t^R$ to compute the $Q, K, V$ matrices, and apply the self-attention mechanism via the following formula:
\begin{equation}
\begin{aligned}
\hat{Z}_t^R = ShiftBack(Att(Q,K,V)),
\end{aligned}
\label{eq:7}
\end{equation}
where $ShiftBack$ means that patches or tokens from different frames are shifted back to their original positions. Since $Z_t^R$ contains patches/tokens from neighboring frames, the standard self-attention mechanism naturally transforms into a spatial-temporal self-attention mechanism. 
Similarly, using TPS encoder $f_3(\cdot)$, we can obtain the IR modal features $\hat{Z}_t^I = f_3(\tilde{I}_t)$, $t = 1,\cdots, T$. In addition, since attention mechanisms are inherently incorporated into the subsequent SD module, we can integrate the attention operation from TPS with the SD module, thereby achieving further reductions in both parameter count and computational complexity.

\subsection{Semantic Diffusion Module}
Despite the effective extraction of sequence-level spatial-temporal features by the STFL module, the RGB and IR modalities still exhibit significant semantic discrepancies due to their inherent imaging characteristics. 
These modality-specific semantic biases hinder the learning of modal-invariant representations, which are essential for cross-modal matching in VVI-ReID. To mitigate this issue, we propose the semantic diffusion module, which leverages the modality-shared text semantics from the CLIP model to bridge the semantic gap between RGB and IR features, and diffuses such shared semantics into the modality-specific features to generate preliminary modal-invariant representations.

\textbf{Text Semantic Extraction} \ \ CLIP's text encoder is pre-trained on a large-scale image-text dataset, enabling it to learn universal semantic representations that are consistent across visual modalities. We first construct a set of identity-related text/language prompts for pedestrian ReID, formulated as: ``A $[X]_1 [X]_2 $ \dots $ [X]_m$ person observed in both day and night conditions.'' where $[X]_i$ denotes a learnable prompt embedding. Then, these prompts are fed into frozen text encoder $f_{\text{text}}(\cdot)$ to obtain the modality-shared text semantic embedding $T_e \in \mathbb{R}^{D}$, where $D$ denotes the feature dimension consistent with the output of the STG encoder. 

\textbf{Cross-Modal Semantic Diffusion} \ \ For the RGB sequence-level features $\{ \hat{Z}_t^R \}_{t=1}^T$ and IR sequence-level features $\{ \hat{Z}_t^I \}_{t=1}^T$ output by the STFL module, we design a semantic diffusion mechanism to inject the shared text semantic embedding $T_e$ into each modality’s features. Specifically, for the RGB modality branch, the $T_e$ is first used to generate the query vector $q_t$ for the $t$-th frame:
\begin{equation}
\begin{aligned}
q_t = W_{query} \cdot T_e.
\end{aligned}
\label{eq:8}
\end{equation}
Then, $\hat{Z}_t^R$ is used to generate the key and value matrices:
\begin{equation}
\begin{aligned}
K_t = W_{key} \cdot \hat{Z}_t^R, \ \
V_t = W_{value} \cdot \hat{Z}_t^R. \\
\end{aligned}
\label{eq:9}
\end{equation}
Finally, the preliminary modal-invariant representation for RGB modality is computed as
\begin{equation}
\begin{aligned}
& \bar{z}_t^R = Att(q_t, K_t, V_t) = V_t \cdot Softmax(\frac{K_t^T \cdot q_t}{\sqrt{d}}), \\
& \bar{Z}_t^R = Concat(\bar{z}_t^R, \hat{Z}_t^R), \ t=1, \cdots, T.
\end{aligned}
\label{eq:10}
\end{equation}
Likewise, for the IR modality branch, $T_e$ is used to generate the query vector, $\hat{Z}_t^I$ is used to produce the key and value matrices, and the cross-modal attention mechanism is utilized to obtain the preliminary modal-invariant representations $\{ \bar{Z}_t^I \}_{t=1}^T$ for IR modality.

By interacting the features of the two modalities with semantic priors in the shared semantic space, the SD module enhances the semantic consistency between RGB and IR modalities. Such semantic alignment of RGB and IR features not only enhances the discriminability of pedestrian representations in complex cross-modal scenarios but also markedly elevates the semantic expression capability of the IR modality.

\subsection{Cross-Modal Interaction Module}
Although the SD module generates preliminary modal-invariant features, the residual modality gap between RGB and IR representations still limits the accuracy of cross-modal matching. To address this, we propose the cross-modal interaction module, which enables fine-grained feature interaction between RGB and IR modalities, further eliminating modality-specific biases and refining the final sequence-level modal-invariant representations.

The CMI module leverages a cross-modal self-attention mechanism to facilitate bidirectional feature interaction between the RGB and IR branches. As illustrated in Fig. \ref{fig:frame}, for the RGB modality branch, the $\bar{Z}_t^R$ is used to calculate the key and value matrices, and the $\bar{Z}_t^I$ is used to calculate the query matrix:
\begin{equation}
\begin{aligned}
K_t = W_{key} \cdot \bar{Z}_t^R, \ \
V_t = W_{value} \cdot \bar{Z}_t^R. \ \
Q_t = W_{query} \cdot \bar{Z}_t^I.
\end{aligned}
\label{eq:11}
\end{equation}
Then, the modal-invariant features for RGB modality are calculated as 
\begin{equation}
\begin{aligned}
F_t^R =& Att(Q_t, K_t, V_t) \\
=& V_t \cdot Softmax(\frac{K_t^T \cdot Q_t}{\sqrt{d}}), \ t=1,\cdots,T.
\end{aligned}
\label{eq:12}
\end{equation}
In the same vein, for the IR modality branch, the $\bar{Z}_t^R$ is used to calculate the query matrix, the $\bar{Z}_t^I$ is used to calculate the key and value matrices, and the modal-invariant features for IR modality $\{F_t^I\}_{t=1}^T$ can be obtained through Eq. (\ref{eq:12}).

Finally, temporal average pooling (TAP) is applied to the $[cls]$ tokens of all modal-invariant features to obtain the sequence-level modal-invariant features $f^R$ and $f^I$, i.e.,
\begin{equation}
\begin{aligned}
& f^R = TAP([f_1^{R,cls}, f_2^{R,cls}, \cdots, f_T^{R,cls}]), \\
& f^I = TAP([f_1^{I,cls}, f_2^{I,cls}, \cdots, f_T^{I,cls}]),
\end{aligned}
\label{eq:13}
\end{equation}
where $f_t^{R,cls}$, $f_t^{I,cls}$ are $[cls]$ tokens of $F_t^R$ and $F_t^I$, respectively.

The CMI module enables targeted bidirectional information exchange between RGB and IR features, allowing each modality to absorb identity-discriminative patterns from the other while suppressing modality-specific noise. This design effectively bridges the residual modality gap that remains after the SD module, making the features more robust to cross-modal discrepancies like texture and thermal pattern differences. Furthermore, by refining features through bidirectional cross-modal attention interactions and temporal average pooling, the CMI module ensures the final sequence-level representations $f^R$ and $f^I$ are not only modal-invariant but also highly discriminative for pedestrian identity. After training, the CMI module can capture discriminative modal-invariant features even when single-modal data is input during the inference phase.

\subsection{Loss Function}
To train our model end-to-end and ensure the learning of discriminative, modal-invariant pedestrian representations, we design a multi-loss system that operates at different stages of the pipeline. Each loss targets specific objectives, from intra-modal feature discrimination to cross-modal alignment.

\textbf{Losses in STFL Module} \ \ 
To enhance the discriminative capability of the encoders in STFL module, we introduce both identity (ID) loss and weighted regularized triplet (WRT) loss during the training of the STG encoder and TPS encoder. The ID loss supervises the model to distinguish different person identities, enabling class-level discriminative learning. The weighted triplet loss minimizes the distance between samples of the same identity while maximizing the distance between samples of different identities, thereby improving intra-class compactness and inter-class separability in the feature space.

For the outputs of STG encoder $\{ \tilde{R}_t, \tilde{I}_t \}_{t=1}^T$, TAP is first applied to each frame's \emph{[cls]} token to obtain sequence-level features, i.e., $f_{\text{STG}}^M = TAP(\tilde{R}_t)$ or $TAP(\tilde{I}_t)$, where $M \in \{R, I\}$ represents different modality features. Then, the ID loss can be defined as:
\begin{equation}
\begin{aligned}
 L_\text{id}^{\text{STG}, M} = -\frac{1}{n_b} \sum_{i=1}^{n_b} q_i \log \big( W_\text{id}^{\text{STG}}(f_{\text{STG},i}^M) \big),
\end{aligned}
\label{eq:14}
\end{equation}
where $ n_b $  is the batch size, $ q_i $  is the ground-truth identity label of the $i$-th sample, $f_{\text{STG},i}^M$ denotes the feature representation of the $i$-th sample extracted from $M$ modality, and $ W_\text{id}^{\text{STG}} $ is the linear classification layer. The WRT loss is defined as:
\begin{equation}
\begin{aligned}
 L_\text{wrt}^{\text{STG}, M} = \frac{1}{n_b} \sum_{i=1}^{n_b} \log \Big( 1 + \exp \Big( \sum_{j} w_{i,j}^{p,M} d_{i,j}^{p,M} \\
 - \sum_{k} w_{i,k}^{n,M} d_{i,k}^{n,M} \Big) \Big),
\end{aligned}
\label{eq:15}
\end{equation}
where $  d_{i,j}^{p,M} $  and $  d_{i,k}^{n,M} $  represent the Euclidean distances between the $i$-th sample and its positive and negative samples, respectively, $w_{i,j}^{p,M} $  and $  w_{i,k}^{n,M} $  are adaptive weight coefficients, calculated as:
\begin{equation}
\begin{aligned}
   &w_{i,j}^p = \frac{\exp(d_{i,j}^{p,M})}{\sum_{d_{i,j} \in P_i} \exp(d_{i,j}^{p,M})}, 
   &w_{i,k}^n = \frac{\exp(-d_{i,k}^{n,M})}{\sum_{d_{i,k} \in N_i} \exp(-d_{i,k}^{n,M})},
\end{aligned}
\label{eq:16}
\end{equation}
where $ P_i $ and $  N_i $  denote the sets of positive and negative samples for the $i$-th sample, respectively. Using the same approach as above and substituting Eq. (\ref{eq:14})–(\ref{eq:16}), we obtain the loss function $ L_\text{id}^{\text{TPS}}$,  $L_\text{wrt}^{\text{TPS}}$ corresponding to the TPS encoder.

Text semantics form the foundation of the SD and CMI modules, to acquire shared text semantics and achieve alignment between text features and RGB/IR modal features, we design the video-to-text (V2T) contrastive loss subsequent to the STG encoder. Specifically, given the sequence-level features $f_{\text{STG}}^R$ and $f_{\text{STG}}^I$ output by the STG encoder and the text feature $T_e$ output by the text encoder, we compute their similarity value as:
\begin{equation}
s = f_v^T W_{\text{proj}} T_e,
\end{equation}
where $f_v = Concat(f_{\text{STG}}^R, f_{\text{STG}}^I)$ denotes the fused video feature, $  W_{\text{proj}}  $  is a learnable projection matrix that maps the video feature into the shared semantic space of text features.  
The similarity value is then passed to the following V2T loss:
\begin{equation}
L_{\text{v2t}} = -\frac{1}{n_b} \sum_{i=1}^{n_b} \log \big( \text{Softmax}(s_i) \big).
\end{equation}
where $ s_i $ denotes the similarity value for the $i$-th sample pair, and \(\text{Softmax}(\cdot)\) operates along the batch dimension.  

Finally, the overall optimization objective of STFL module can be formulated as:
\begin{equation}
  L_\text{STFL} = L_\text{id}^{\text{STG}} + L_\text{wrt}^{\text{STG}} + L_\text{id}^{\text{TPS}} + L_\text{wrt}^{\text{TPS}} + \lambda_1 L_{\text{v2t}},
\end{equation}
where $\lambda_1$ is a hyper-parameter.

\textbf{Losses in CMI Module} \ \ 
The previously introduced ID loss and WRT loss are effective in improving feature discriminability, yet they lack the capability to facilitate the learning of discriminative and modal-invariant features. To tackle this limitation, we integrate the modality-discriminative (MD) loss \citep{Feng2023MDloss} and modality-shared enhancement loss (MSEL) \citep{Lu2023MSELloss} subsequent to the CMI module (as illustrated in Fig. \ref{fig:motivation}). Among these, the MD loss enforces intra-class and inter-modality feature compactness while enhancing inter-class and inter-modality feature separability. In contrast, the MSEL loss suppresses spurious modal-specific features and strengthens robust modal-invariant representations.

Formally, given a pair of sequence-level modal-invariant features $f^R$ and $f^I$ output by CMI, we define the MD loss as
\begin{equation}
\begin{aligned}
L_{\text{MD}} = - \text{log} \frac{\text{exp}(\text{sim}(f^R_i, f^I_i)/\tau)}{\sum \limits_{i \neq j} \text{exp}(\text{sim}(f^R_i, f^I_j)/\tau)},
\end{aligned}
\label{eq:20}
\end{equation}
where $f^R_i, f^I_i$ denotes the feature representations of the $i$-th sample of RGB modality and IR modality, respectively. 

Assume that there are $n_b$ RGB-IR video sample pairs in each batch, which are derived from $P$ pedestrians, and each pedestrian has $K$ sample pairs, i.e., $n_b = PK$. For the $i$-th sample pair, its intra-modality average distance to the samples of the same category can be calculated as: 
\begin{equation}
\begin{aligned}
D_{i}^{1,I} = \frac{1}{K-1} \underset {i \neq j}{\sum \limits_{i=1}^{K-1}} D(f_{i}^I, f_{j}^I), \ D_{i}^{1,R} = \frac{1}{K-1} \underset {i \neq j}{\sum \limits_{i=1}^{K-1}} D(f_{i}^R, f_{j}^R).
\end{aligned}
\label{eq:21}
\end{equation}
And its cross-modality average distance to the samples of the same category can be calculated as: 
\begin{equation}
\begin{aligned}
D_{i}^{2,I} = \frac{1}{K} \sum \limits_{i=1}^{K} D(f_{i}^I, f_{j}^R), \ D_{i}^{2,R} = \frac{1}{K} \sum \limits_{i=1}^{K} D(f_{i}^R, f_{j}^I).
\end{aligned}
\label{eq:22}
\end{equation}
Then, the MSEL loss can be defined as
\begin{equation}
\begin{aligned}
L_{\text{MSEL}} = \frac{1}{2PK} \sum_{p=1}^{P} \bigg[ \sum_{i=1}^{K} \bigg( \big(D_{i}^{1,I}-D_{i}^{2,I}\big)^2 + \big(D_{i}^{1,R} \\
- D_{i}^{2,R}\big)^2  \bigg) \bigg].
\end{aligned}
\label{eq:23}
\end{equation}
When \(D^{1,I}\) is equal to \(D^{2,I}\) and \(D^{1,R}\) is equal to \(D^{2,R}\), the loss function converges to its optimal value. At this point, the two types of modal-invariant features of the same sample should exhibit a high degree of similarity. In other words, the objective of the loss function is to make the two interacted modal-invariant features as close as possible to each other.

\textbf{Overall Loss Function} \ \ 
Finally, the overall loss is defined as:
\begin{equation}
L = L_\text{STFL} + \lambda_2 L_{\text{MSEL}} + \lambda_3 L_{\text{MD}},
\end{equation}
where $\lambda_2$ and $\lambda_3$  are hyper-parameters used to balance the loss terms.

\begin{table*}[!t]
	\caption{Comparison of CMC (\%) and mAP (\%) performances with the state-of-the-art methods on HITSZ-VCM dataset. }
	\label{result vcm}
	\centering 
\renewcommand{\arraystretch}{1}
\footnotesize
		\begin{tabular}{l c c c c c c  c c c c c c c}
			\hline
			\multirow{2.5}{*}{Method} &\multirow{2.5}{*}{Venue}&\multirow{2.5}{*}{Seq-length} & \multicolumn{4}{c}{Infrared to Visible} & \multicolumn{4}{c}{Visible to Infrared} & \multirow{2.5}{*}{Average} \\ 
\cmidrule(l){4-7}  \cmidrule(l){8-11}
			& & & Rank-1& Rank-5& Rank-10& mAP& Rank-1& Rank-5& Rank-10 & mAP \\ 
            \hline
			DDAG\citep{ye2020dynamic}    &ECCV'20     &6 & 54.6 &69.8 &76.1 &39.3 &59.0 &74.6 &79.5 &41.5 & 61.5 \\
            Lba\citep{park2021learning}  &ICCV'21     &6 & 46.4 &65.3 &72.2 & 31.0&49.3 &69.3 &75.9 & 32.4 & 55.2 \\
            MPANet\citep{wu2021discover} &CVPR'21     &6 & 46.5 &63.1 &70.5 &35.3 &50.3 &67.3 &73.6 &37.8 & 55.5 \\
            VSD\citep{tian2021farewell}  &CVPR'21     &6 & 54.5 &70.0 &76.3 &41.2 &57.5 &73.7 &79.4 &43.5 & 63.3 \\
            CAJL\citep{ye2021channel}    &ICCV'21     &6 & 56.6 &73.5 &79.5 &41.5 &60.1 &74.6 &79.9 &42.8 & 63.6 \\
            DEEN\citep{zhang2023diverse} &CVPR'23     &6 & 53.7 &74.8 &80.7 &50.4 &49.8 &71.6 &81.0 &48.6 & 63.7\\ 
            CLIP-ReID\citep{li2023clip}  &AAAI'23     &6 & 58.4 &73.2 &79.8 &45.3 &60.4 &76.9 &83.7 &43.5 & 65.2 \\
            UCT\citep{yuan2024unbiased}  &    TOMM'24 &6 & 58.4 &76.8 &80.2 &43.2 &61.2 &75.8 &81.5 &49.0 & 65.8 \\
			HOS-Net\citep{qiu2024high}   &AAAI'24     &6 & 61.3 &75.1 &79.0 &46.0 &63.9 &74.6 &81.4 &47.9 & 66.2 \\   
            \hline  
            TF-CLIP\citep{yu2024tf}      &AAAI'24     &6 & 62.3 &76.2 &81.6 &47.5 &62.2 &79.6 &85.5 &45.5 & 67.6 \\                                                                                                                                                                      
            MITML\citep{lin2022learning} &CVPR'22     &6 & 63.7 &76.9 &81.7 &45.3 &64.5 &79.0 &83.0 &47.7 & 67.7 \\
            SAADG\citep{zhou2023video}   &ACM MM'23   &6 & 69.2 &80.6 &85.0 &53.8 &73.1 &83.5 &86.9 &56.1 & 73.5 \\
            IBAN\citep{Li2023IBAN}       &TCSVT'23    &6 & 65.0 &78.3 &83.0 &48.8 &69.6 &81.5 &85.4 &51.0 & 70.3 \\
            AuxNet\citep{du2023video}    &TIFS'23     &6 & 51.1 & -   & -   & -   &46.0 &54.6 &-    &48.7 & 50.1 \\
            CST\citep{feng2024cross}     &TMM'24      &6 & 69.4 &81.1 &85.8 &51.2 &72.6 &83.4 &86.7 &53.0 & 72.9 \\ 
            DIRL\citep{Wang2025DIRL}     &TOMM'25     &6 & 65.2 &79.1 &84.6 &47.9 &67.0 &81.7 &84.2 &50.2 & 70.0 \\ 
            HD-GI \citep{Zhou2025HDGI}&Inf. Fusion'25 &- & 71.4 &81.7 &84.9 &58.0 &75.0 &84.4 &87.3 &60.2 & 75.4 \\ 
            X-ReID \citep{yu2025xreid}&AAAI'26        &10& 73.4 &85.0 &--   &60.5 &76.1 &87.1 &--   &59.6 & 73.6 \\ 
            VLD\citep{li2025video}       &TIFS'25     &6 &\underline{74.3} &\textbf{85.0} &\underline{88.4} &\underline{60.2} &\underline{74.6} &\underline{86.4} &\underline{90.0} &\textbf{58.6} & \underline{77.2} \\
            \hline
			Ours&- &6&\textbf{75.1}&\underline{84.6}&\textbf{89.0}&\textbf{60.9}&\textbf{75.2}&\textbf{87.4}&\textbf{91.5} &\underline{58.2} & \textbf{77.7} \\ 
            \hline
	\end{tabular} 
\end{table*}
\section{Experiment} \label{sec:IV}

\subsection{Datasets and Evaluation Metrics}
To rigorously demonstrate the effectiveness of the proposed method, we conduct extensive experiments on two large-scale VVI-ReID datasets: HITSZ-VCM\citep{lin2022learning} dataset and BUPTCampus\citep{du2023video} dataset. These benchmarks are widely recognized in the community and provide diverse scenarios for validating both cross-modal feature learning and temporal modeling.

{\textbf{HITSZ-VCM Dataset\citep{lin2022learning}} was collected using six visible and six infrared cameras, covering 927 identities with 251,452 RGB images and 211,807 infrared images. The data are organized into video sequences, each consisting of 24 consecutive frames. The dataset is partitioned into 500 identities for training and 427 identities for testing. As the first large-scale benchmark dedicated to VVI-ReID, HITSZ-VCM serves as a representative platform for evaluating temporal modeling in cross-modal scenarios.

\textbf{BUPTCampus Dataset\citep{du2023video}} was captured using six binocular RGB–IR cameras, comprising 3,080 identities, 16,826 video sequences, and a total of 1,869,366 images. It is divided into three subsets: 1,074 identities for primary learning, 930 identities for auxiliary learning, and 1,076 identities for testing. Compared with HITSZ-VCM, BUPTCampus offers a much larger scale and provides nearly pixel-aligned RGB–IR video samples, which facilitate the investigation of cross-modal representation learning and auxiliary training strategies.

\textbf{Evaluation Metrics.} For both datasets, performance is measured using the cumulative matching characteristic (CMC) curve and mean average precision (mAP). To comprehensively assess cross-modal retrieval, two evaluation settings are adopted: infrared-to-visible, where infrared tracklets are used as queries against visible tracklets, and visible-to-infrared, where visible tracklets are used as queries against infrared tracklets. These complementary protocols ensure a fair and rigorous validation of the proposed approach.

\subsection{Implementation Details}
The proposed LSMRL method was implemented using the PyTorch on a single A800 GPU. To ensure the reproducibility of the experiments, all experiments were conducted with a fixed random seed (set to 42). The proposed method leverages the CLIP vision encoder (ViT-B-16) alongside its text encoder to extract vision and text features. The input video frames were resized to $288 \times 144$ pixels, and standard data augmentation techniques were employed on them. The model was trained using the Adam optimizer with a learning rate of $2.5 \times 10^{-5}$, and a cosine learning rate decay strategy was employed to dynamically adjust the learning rate. The training epoch and batch size were set to 60 and 32, respectively. Each batch contains video sequences of 2 modalities, with 4 pedestrians per modality and 4 video sequences corresponding to each pedestrian. And 6 frames are randomly sampled from each sequence. The hyper-parameters $\lambda_1$, $\lambda_2$ and $\lambda_3$ set to 0.1, 0.05 and 0.5, respectively. The temporal heads $k$ is set to 6. Notably, during the inference phase, we will remove the SD module and text encoder to improve inference speed.

\begin{table*}[!t]
	\caption{Comparison of CMC (\%) and mAP (\%) performances with the state-of-the-art methods on BUPTCampus dataset.}
	\label{result bupt}
	\centering
	\renewcommand{\arraystretch}{1}
\footnotesize
		\begin{tabular}{lccccccccccc}
			\hline
			\multirow{2.5}{*}{Method} &\multirow{2.5}{*}{Venue}&\multirow{2.5}{*}{Seq-length} & \multicolumn{4}{c}{Infrared to Visible} & \multicolumn{4}{c}{Visible to Infrared} & \multirow{2.5}{*}{Average} \\ 
\cmidrule(l){4-7}  \cmidrule(l){8-11}
			& & & Rank-1& Rank-5& Rank-10& mAP& Rank-1& Rank-5& Rank-10 & mAP \\ 
\hline
			DDAG\citep{ye2020dynamic}    &ECCV'20  &10&46.3 &68.2 &74.4 &43.1 &40.4 &40.9 &61.4 &58.5 &54.2   \\
			Lba\citep{park2021learning}  &ICCV'21  &10&39.1 &58.7 &66.5 &37.1 &32.1 &54.9 &65.1 &32.9 & 48.3 \\
            AGW\citep{ye2021deep}        &TPAMI'21 &10&43.7 &64.4 &73.2 &41.1 &36.4 &60.1 &67.2 &37.4 & 52.9 \\
            MMN\citep{zhang2021towards}  &ACM MM'21&10&43.7 &65.2 &73.5 &42.8 &40.9 &67.2 &74.4 &41.7 & 56.2 \\
			CAJL\citep{ye2021channel}    &ICCV'21  &10&45.0 &70.0 &77.0 &43.6 &40.5 &66.8 &73.3 &41.5 & 57.2 \\
			DART\citep{yang2022learning} &CVPR'22  &10&53.3 &75.2 &81.7 &50.5 &52.4 &70.5 &77.8 &49.1 & 63.8 \\
			DEEN\citep{zhang2023diverse} &CVPR'23  &10&53.7 &74.8 &80.7 &50.4 &49.8 &71.6 &81.0 &48.6 & 63.8 \\
            UCT\citep{yuan2024unbiased}  &TOMM'24  &10&56.5 &75.2 &83.2 &56.0 &56.0 &78.0 &83.7 &53.7 & 67.8 \\
			HOS-Net\citep{qiu2024high}   &AAAI'24  &10&54.9 &74.2 &83.4 &55.2 &53.1 &75.8 &81.0 &50.8 & 66.1 \\
            SIMFGA\citep{zuo2025spatio}  &IMAVIS'25&10&55.1 &73.8 &80.5 &54.6 &54.9 &76.6 &82.6 &51.8 & 66.2 \\ 
            AuxNet\citep{du2023video}    &TIFS'23  &10&65.2 &81.8 &86.1 &62.2 &66.5 &83.1 &\underline{87.9} &64.1 & 74.6 \\ 
            DIRL\citep{Wang2025DIRL}     &TOMM'25  &8 &67.6 &83.2 &\underline{87.5} &63.4 &67.2 &83.0 &87.4 &65.3 & \underline{75.6} \\
            X-ReID \citep{yu2025xreid}   &AAAI'26  &10&\underline{68.2} &\underline{88.4} &--   &\underline{68.5} &\underline{68.8} &\underline{84.8} &--   &\underline{65.9} & 74.1 \\ 
\hline
            CLIP-ReID\citep{li2023clip}  &AAAI'23  &6 &49.0 &73.0 &81.2 &50.4 &51.0 &75.4 &80.0 &49.8 & 62.5 \\
            TF-CLIP\citep{yu2024tf}      &AAAI'24  &6 &49.4 &76.8 &83.7 &51.9 &52.5 &75.2 &81.5 &51.8 & 65.4 \\
            MITML\citep{lin2022learning} &CVPR'22  &6 &50.2 &68.3 &75.7 &46.3 &49.1 &68.0 &75.4 &47.5 & 60.1 \\
            VLD\citep{li2025video}       &TIFS'25  &6 &\underline{65.3} &\underline{84.9} &\underline{89.7} &\underline{63.5} &\underline{65.8} &\underline{83.0} &\underline{87.9} &\underline{63.0} & \underline{75.4} \\
\hline
			Ours&-&6 &\textbf{71.1}&\textbf{86.0}&\textbf{89.1}&\textbf{67.5}&\textbf{67.0}&\textbf{84.4}&\textbf{89.1}&\textbf{65.7} & \textbf{77.5} \\
			\hline
	\end{tabular}
\end{table*}

\subsection{Experimental Results}
 In order to demonstrate the effectiveness and competitiveness of our approach, we present a comprehensive comparison of LSMRL against SOTA methods on two widely recognized VVI-ReID benchmarks. Specifically, the comparison methods includes the IVI-ReID category (the upper of Tab. \ref{result vcm}), as well as the VVI-ReID category (the bottom of the Tab. \ref{result vcm}).

\subsubsection{\textbf{Comparative experiments on the HITSZ-VCM dataset}}
The proposed method was first evaluated on the HITSZ-VCM benchmark, which demonstrated significant performance improvements over state-of-the-art methods in both rank accuracy and mAP, as shown in Tab. \ref{result vcm}. More specifically, compared with the latest CLIP-based VLD method, the proposed method achieves significant improvements in key metrics: in the Infrared to Visible (I2V) task, the proposed method outperforms VLD in Rank-1 (75.1\% vs. 74.3\%), Rank-10 (89.0\% vs. 88.4\%), and mAP (60.9\% vs. 60.2\%); in the Visible to Infrared (V2I) task, the proposed method shows superiority in Rank-1 (75.2\% vs. 74.6\%), Rank-5 (87.4\% vs. 86.4\%), and Rank-10 (91.5\% vs. 90.0\%), while its mAP is comparable to that of VLD. This result demonstrates that by means of targeted cross-modal interaction and spatial-temporal feature enhancement, the proposed method can further tap into the cross-modal video alignment potential of CLIP, thus verifying the effectiveness of the method.

Compared with traditional IVI-ReID methods, the proposed method achieves significant improvements in core metrics: for example, in the I2V task the proposed method outperforms HOS-Net by 8.3, 5.2, 4.8, and 9.7 percentage points, which verifies the importance of temporal feature modeling. When compared with VVI-ReID methods (e.g., CST, SAADG, etc.), the proposed method also maintains leading performance: for instance, in the I2V task, the proposed method outperforms SAADG by 5.9, 4.0, 4.0, and 7.1 percentage points in Rank-1, Rank-5, Rank-10, and mAP, respectively. This indicates that the proposed method is more competitive in spatial-temporal modeling of video sequence and cross-modal matching. Furthermore, a comparison between CLIP-based methods (including VLD, X-ReID and the proposed method) and other VVI-ReID methods reveals that CLIP-based methods generally achieve superior performance across all metrics. This indicates that the semantic generalization capability of CLIP pre-trained features is conducive to improving the accuracy of cross-modal retrieval.

\subsubsection{\textbf{Comparative experiments on the BUPTCampus dataset}}
To further evaluate scalability, we also conducted comparative experiments on a larger and more diverse VVI-ReID dataset, and the experimental results are shown in Tab. \ref{result bupt}. From this table, we observe that the proposed method significantly outperforms state-of-the-art methods, even with a sequence length of 6. More specifically, compared with 6-frame SOTA VLD, our approach achieves significant performance advantages: in the I2V task, it surpasses VLD by 5.8, 1.1, and 4.0 percentage points in Rank-1, Rank-5, and mAP, respectively, while its Rank-10 is comparable to that of VLD; in the V2I task, it outperforms VLD by 1.2, 1.4, 1.2, and 2.7 percentage points in Rank-1, Rank-5, Rank-10, and mAP, respectively. Even when compared with the 10-frame SOTA method X-CLIP, our approach remains competitive in terms of Rank-1 accuracy and mAP. This can be attributed to the synergistic effect of our proposed spatial-temporal feature learning and cross-modal interaction mechanisms, which fully exploit CLIP's pre-trained semantic generalization capabilities while effectively mitigating the modality gap and capturing fine-grained temporal dynamics in video sequences. Additionally, the multi-loss optimization strategy enhances the discriminability and modal invariance of features, enabling the model to maintain superior performance even on large-scale diverse datasets with a shorter sequence length.

Compared with IVI-ReID methods (e.g., UCT), our method achieves substantial performance improvements across both I2V and V2I tasks. For instance, in the I2V task, our method outperforms UCT by 14.6 percentage points in Rank-1 (71.1\% vs. 56.5\%), 10.8 percentage points in Rank-5 (86.0\% vs. 75.2\%), 5.9 percentage points in Rank-10 (89.1\% vs. 83.2\%), and 11.5 percentage points in mAP (67.5\% vs. 56.0\%). This reconfirms the importance of temporal feature modeling in VVI-ReID tasks. Furthermore, we find again that compared with traditional VVI-ReID methods, CLIP-based approaches achieve superior performance, which can be attributed to the fact that the pre-trained CLIP model has strong cross-modal semantic alignment capabilities and broad domain generalization.

\subsection{Ablation Studies}
\begin{table}[!t]
	\caption{The effects of our proposed components performed on BUPTCampus dataset, where ``R@1'', ``R@5'' and ``R@10'' denote Rank-1, Rank-5 and Rank-10, respectively.}
	\label{component_ablation}
	\centering
\scriptsize 
\setlength{\tabcolsep}{2.5pt}  
		\begin{tabular}{ccccccccccccc}
			\hline
			\multirow{2}{*}{Base} & \multicolumn{2}{c}{STFL} & \multirow{2}{*}{SD} & \multirow{2}{*}{CMI} & \multicolumn{4}{c}{Infrared to Visible} & \multicolumn{4}{c}{Visible to Infrared} \\
			\cline{2-3} \cline{6-13}  \cline{10-13}
			& STG & TPS & & & R@1 & R@5 & R@10 & mAP & R@1 & R@5 & R@10 & mAP \\
			\hline
			\checkmark & - & - & - & - & 45.8 & 74.0 & 82.4 & 48.4 & 45.3 & 69.5 & 77.3 & 46.7 \\
			\checkmark & \checkmark & - & - & - & 48.2 & 74.8 & 83.4 & 49.6 & 46.5 & 70.4 & 78.1 & 47.3 \\
			\checkmark & - & \checkmark & - & - & 52.7 & 75.7 & 84.8 & 53.2 & 51.8 & 75.2 & 80.7 & 51.6 \\
			\checkmark & \checkmark & \checkmark & - & - & 53.1 & 77.2 & 85.2 & 54.2 & 53.6 & 76.6 & 82.4 & 53.8 \\
			\checkmark & \checkmark & \checkmark & \checkmark & - & 66.9 & 83.2 & 86.8 & 62.5 & 65.8 & 81.9 & 84.5 & 61.8 \\
			\checkmark & \checkmark & \checkmark & \checkmark & \checkmark & \textbf{71.1} & \textbf{86.0} & \textbf{89.1} & \textbf{67.5} & \textbf{67.0} & \textbf{84.4} & \textbf{89.1} & \textbf{65.7} \\
			\hline
		\end{tabular}
\vspace{-8pt} 
\end{table}

\textbf{Ablation on Modules}:
To further validate the effectiveness of each proposed component, we conduct an ablation study on the BUPTCampus dataset, as shown in Tab. \ref{component_ablation}. The analysis is performed by incrementally adding each component of our framework and evaluating the performance in terms of Rank accuracy and mAP. Note that ``Base'' is an abbreviation for ``baseline'', which denotes the direct use of CLIP's pre-trained visual encoder for VVI-ReID.

To evaluate the effectiveness of the STFL module, we integrate it into the ``Base''. As shown in Tab. \ref{component_ablation}, the addition of STFL led to significant performance improvements. In the I2V task, the Rank-1 accuracy increased from 45.8\% to 53.1\% (+7.3\%), and the mAP improved from 48.4\% to 54.2\% (+5.8\%). Similarly, in the V2I task, the Rank-1 accuracy increased from 45.3\% to 53.6\% (+8.3\%), and the mAP increased from 46.7\% to 53.8\% (+7.1\%). Since the STFL module consists of STG encoder and TPS encoder, we also analyze the effectiveness of each component individually, i.e., we integrate them into the ``Base'' respectively. For the STG encoder, in the I2V task, the Rank-1 accuracy increased from 45.8\% to 48.2\% (+2.4\%), and the mAP improved from 48.4\% to 49.6\% (+1.2\%). In the V2I task, the Rank-1 accuracy increased from 45.3\% to 46.5\% (+1.2\%), and the mAP increased from 46.7\% to 48.3\% (+1.6\%). These results confirm that the STG encoder effectively generates robust and discriminative spatial-temporal feature representations. For the TPS encoder, in the I2V task, the Rank-1 accuracy increased from 45.8\% to 52.7\% (+6.9\%), and the mAP improved from 48.4\% to 53.2\% (+4.8\%). In the V2I task, the Rank-1 accuracy increased from 45.3\% to 51.8\% (+6.5\%), and the mAP increased from 46.7\% to 51.6\% (+4.9\%). This indicates that compared with the STG, the TPS can model spatial-temporal features more effectively. Furthermore, we find that using the STG and TPS simultaneously can further improve the experimental results, which indicates that the two modules exhibit a synergistic effect in spatial-temporal feature modeling, which is consistent with our theoretical analysis.

In addition, to evaluate the effectiveness of the SD module, we integrate it into the ``Base+STFL'' configuration for comparative experiments. In the I2V task, the Rank-1 accuracy increased from 53.1\% to 66.9\% (+13.8\%), and the mAP improved from 54.2\% to 62.5\% (+8.3\%). In the V2I task, the Rank-1 accuracy increased from 53.6\% to 65.8\% (+12.2\%) and the mAP increased from 53.8\% to 61.8\% (+8.0\%). These results clearly demonstrate that the SD module plays a crucial role in bridging the semantic gap between RGB and infrared modalities. By diffusing modality-shared text semantics into the feature representations of both modalities, it effectively enhances the consistency of cross-modal features, thereby significantly boosting the model’s retrieval accuracy and robustness in VVI-ReID tasks. 

To evaluate the effectiveness of the CMI module, we integrate it into the ``Base+STFL+SD'' configuration. It can be seen that in the I2V task, the Rank-1 accuracy increased from 66.9\% to 71.1\% (+4.2\%), and the mAP improved from 62.5\% to 67.5\% (+5.0\%). In the V2I task, the Rank-1 accuracy increased from 65.8\% to 67.0\% (+2.2\%), and the mAP increased from 61.8\% to 65.7\% (+3.9\%). These results indicate that the CMI module can effectively enhance cross-modal feature interaction, further narrow the modality gap, and thereby learn modality-invariant features with both strong discriminative power and generalization ability. We guess this is because the CMI module facilitates dynamic information exchange between RGB and IR features through bidirectional semantic communication between them, which compensates for the limitations of the SD module.

\begin{table}[!t]
	\caption{The influences of different losses on BUPTCampus dataset.}
	\label{loss_ablation}
	\centering
\scriptsize 
\setlength{\tabcolsep}{3pt}  
		\begin{tabular}{cccccccccccc}
			\hline
			\multirow{2}{*}{$L_\text{STFL}$} &  \multirow{2}{*}{$L_\text{MSEL}$} & \multirow{2}{*}{$L_\text{MD}$} & \multicolumn{4}{c}{Infrared to Visible} && \multicolumn{4}{c}{Visible to Infrared} \\
			\cline{4-7} \cline{9-12}
			&  &  & R@1 & R@5 & R@10 & mAP && R@1 & R@5 & R@10 & mAP \\
			\hline
\checkmark & -          & -          & 68.2 & 83.1 & 87.9 & 65.2 && 63.8 & 82.0 & 86.7 & 63.1 \\
\checkmark & \checkmark & -          & 69.5 & 85.0 & 88.7 & 66.1 && 65.7 & 83.2 & 88.4 & 64.6 \\
\checkmark & -          & \checkmark & 69.2 & 84.4 & 88.3 & 65.6 && 64.9 & 82.6 & 87.1 & 63.4 \\
\checkmark & \checkmark & \checkmark & \textbf{71.1} & \textbf{86.0} & \textbf{89.1} & \textbf{67.5} && \textbf{67.0} & \textbf{84.4} & \textbf{89.1} & \textbf{65.7} \\
			\hline
\end{tabular}
\vspace{-8pt} 
\end{table}

\textbf{Ablation on Losses}:
To further validate the influence of each loss function, we conduct an ablation study on the BUPTCampus dataset, the results are shown in Tab. \ref{loss_ablation}. Note that, in the first row, we additionally introduce the ID loss after the CMI module to train the latter two modules. As can be seen from this table, the model performs poorly when only identity-level loss (i.e., $L_\text{STFL}$) is used, while it achieves the best performance\,---\,increasing the Rank-1 and mAP from 68.2\% and 65.2\% to 71.1\% and 67.5\% in the I2V task, and from 63.8\% and 63.1\% to 67.0\% and 65.7\% in the V2I task\,---\,when both modality-level losses (i.e., $L_\text{MSEL}$ and $L_\text{MD}$) are enabled. This phenomenon fully validates the core role of modality-level losses in enhancing the learning of modal-invariant features, which is consistent with the motivation of this paper.

\subsection{Visualization}
To comprehensively validate the effectiveness of our method in cross-modal video person re-identification, we conduct a series of visualization analyses from multiple perspectives.

\textbf{Feature Distance Distribution Analysis}:
To verify the discriminative capability of our method in VVI-ReID task, we first analyze the distance distribution of our method and the baseline (i.e., the pre-trained CLIP). Specifically, we compute the intra-class distances of samples belonging to the same identity and the inter-class distances of samples from different identities, and compare the distribution differences between the baseline model and our proposed method, and the experimental results are illustrated in Fig. \ref{Feature Distance}(a)–(b). From these sub-figures we observe that although the baseline can partially separate intra-class distances from inter-class distances, the two distributions still exhibit substantial overlap. In contrast, our method significantly increases the inter-class distance while maintaining the stability of the intra-class distance, thereby yielding a larger mean difference $\delta$ and a clearer decision boundary. These results demonstrate that the proposed method effectively enhances the separability of cross-modal representations.

\textbf{T-SNE Visualization}:
To further illustrate the improvement in feature discriminability brought by our method, we randomly select several identities and employ t-SNE to embed the high-dimensional features into a two-dimensional space for visualization, as shown in Fig. \ref{Feature Distance}(c)–(d).
Fig. \ref{Feature Distance}(c) presents the feature distribution produced by the baseline model. As observed, although the baseline is able to cluster samples from the same modality with the same identity, it fails to clearly separate different identities. Moreover, RGB and IR samples of the same identity are still noticeably misaligned, reflecting insufficient cross-modal feature consistency.
Fig. \ref{Feature Distance}(d) shows the feature distribution generated by our method. It can be seen that our approach substantially improves the aggregation of RGB–IR features from the same identity, leading to highly overlapped cross-modal clusters. This result indicates that we have learned more robust modal-invariant features and directly verifies the effectiveness of the SD and CMI modules. Meanwhile, samples from different identities are more distinctly separated, effectively avoiding class mixing. These visualizations provide strong evidence that the proposed LSMRL markedly enhances the discriminability of cross-modal representations.

\begin{figure}[!t]
\centering
\begin{subfigure}{0.49\linewidth}
  \centering
  \includegraphics[width=\linewidth]{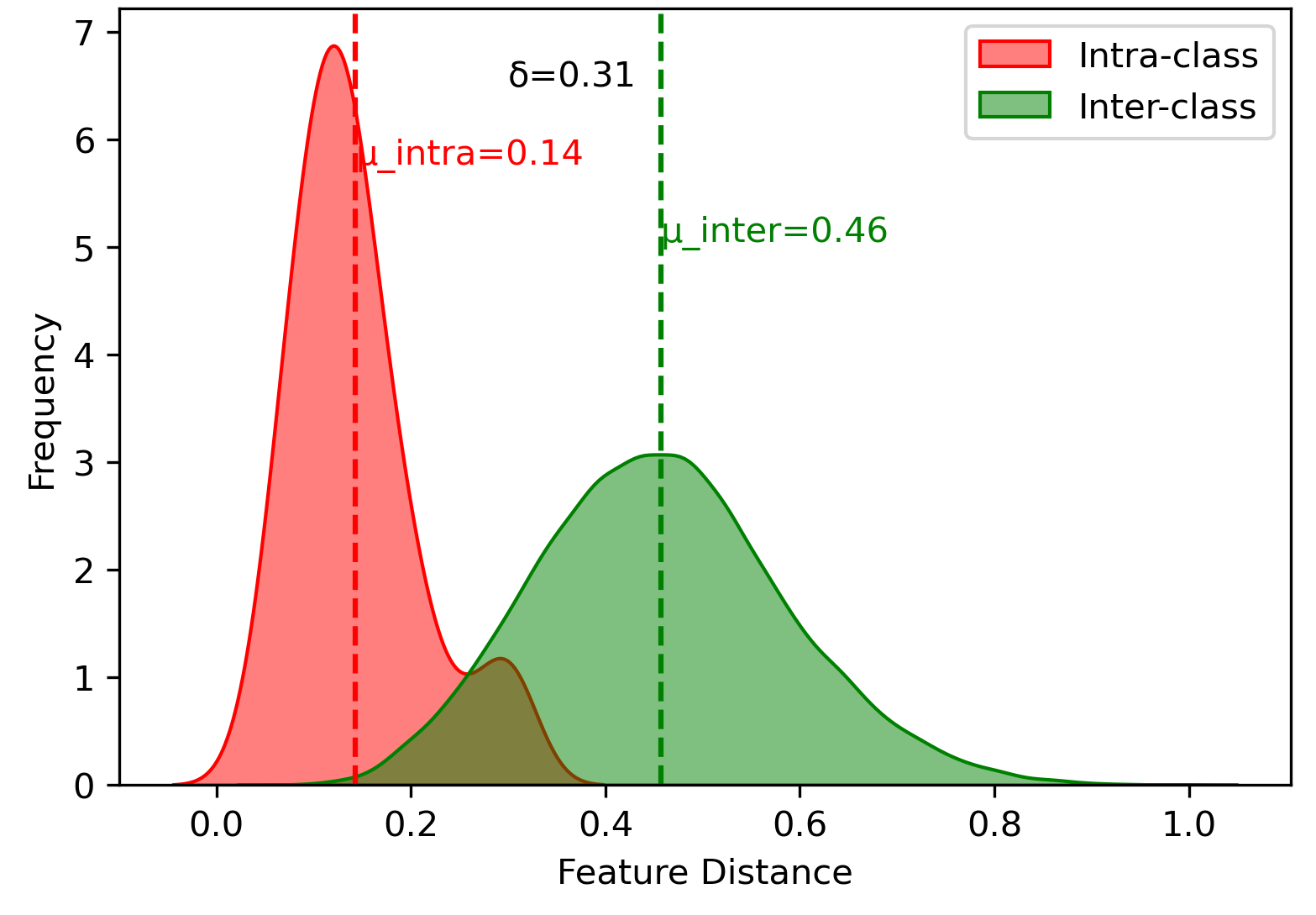}
  \caption{Baseline}
  \label{distance_base}
\end{subfigure}
\hfill
\begin{subfigure}{0.49\linewidth}
  \centering
  \includegraphics[width=\linewidth]{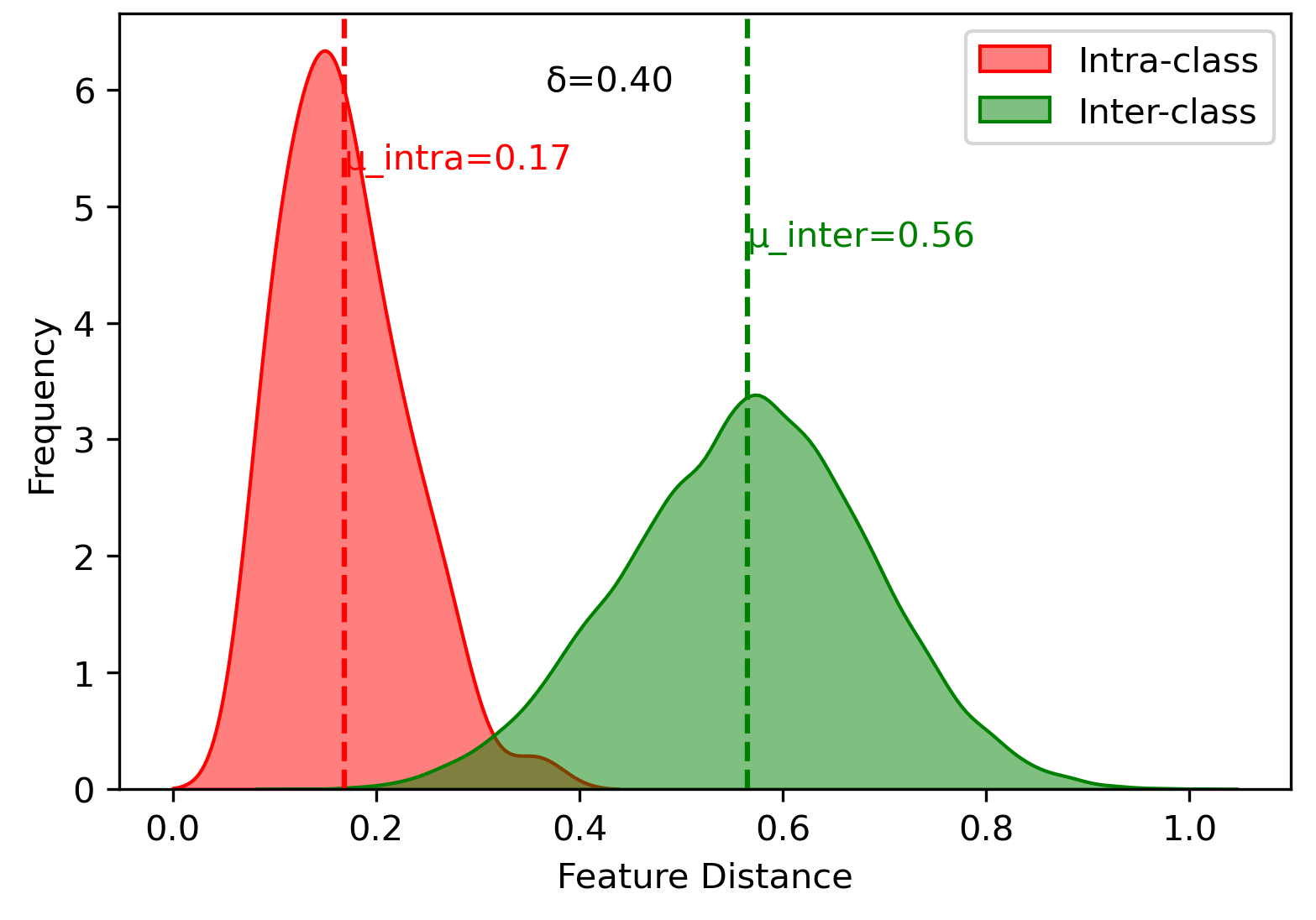}
  \caption{Ours}
  \label{distance_ours}
\end{subfigure}
\hfill
\begin{subfigure}{0.49\linewidth}
  \centering
  \includegraphics[width=\linewidth]{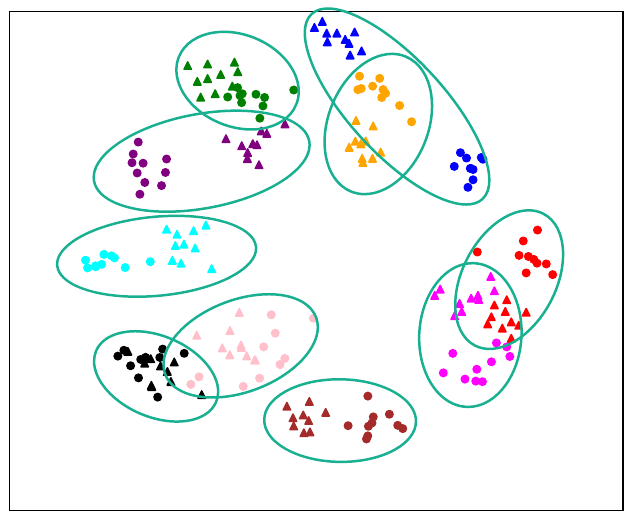}
  \caption{Baseline}
  \label{tsne_base}
\end{subfigure}
\hfill
\begin{subfigure}{0.49\linewidth}
  \centering
  \includegraphics[width=\linewidth]{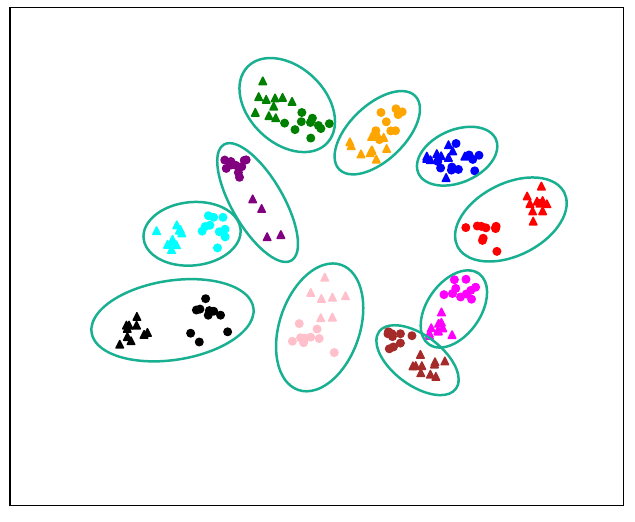}
  \caption{Ours}
  \label{tsne_ours}
\end{subfigure}
\caption{ The distribution of distances and features of the baseline and our approach on the test set of BUPTCampus. (a-b) The distributions of the intra-class distances (in red) and inter-class distances (in green). The larger separation between intra-class and inter-class distances demonstrates improved modality alignment and discrimination. (c-d) Visualization of the corresponding feature space by T-SNE. Each unique color in this visualization represents an identity. Visible and infrared modalities are symbolized as solid circles and triangles, respectively.}
\label{Feature Distance}
\vspace{-8pt}
\end{figure}

\textbf{CAM Visualization}:
Fig. \ref{cam} displays the CAM visualization of focus regions across three methods for both visible and infrared modalities. As can be seen from the figure, the baseline model exhibits disorganized attention distribution, failing to stably lock onto discriminative body parts of pedestrians (e.g., torso and limbs), and thus has significant limitations in focus region modeling. This is because the baseline cannot effectively perceive temporal features, moreover, pre-trained exclusively on visible-light images, it is unable to handle infrared data efficiently. While the VLD model narrows its attention scope to the pedestrian body region, it still suffers from slight attention diffusion, which we conjecture arises from its lack of cross-modal interaction and modality-level loss constraints. Through targeted improvements, the proposed method can tightly focus its attention on core identity feature regions across all frames in both modalities, with uniform and stable distribution, demonstrating enhanced attention consistency and semantic alignment.

\begin{figure}[!t]
\centering
  \includegraphics[width=0.5\textwidth]{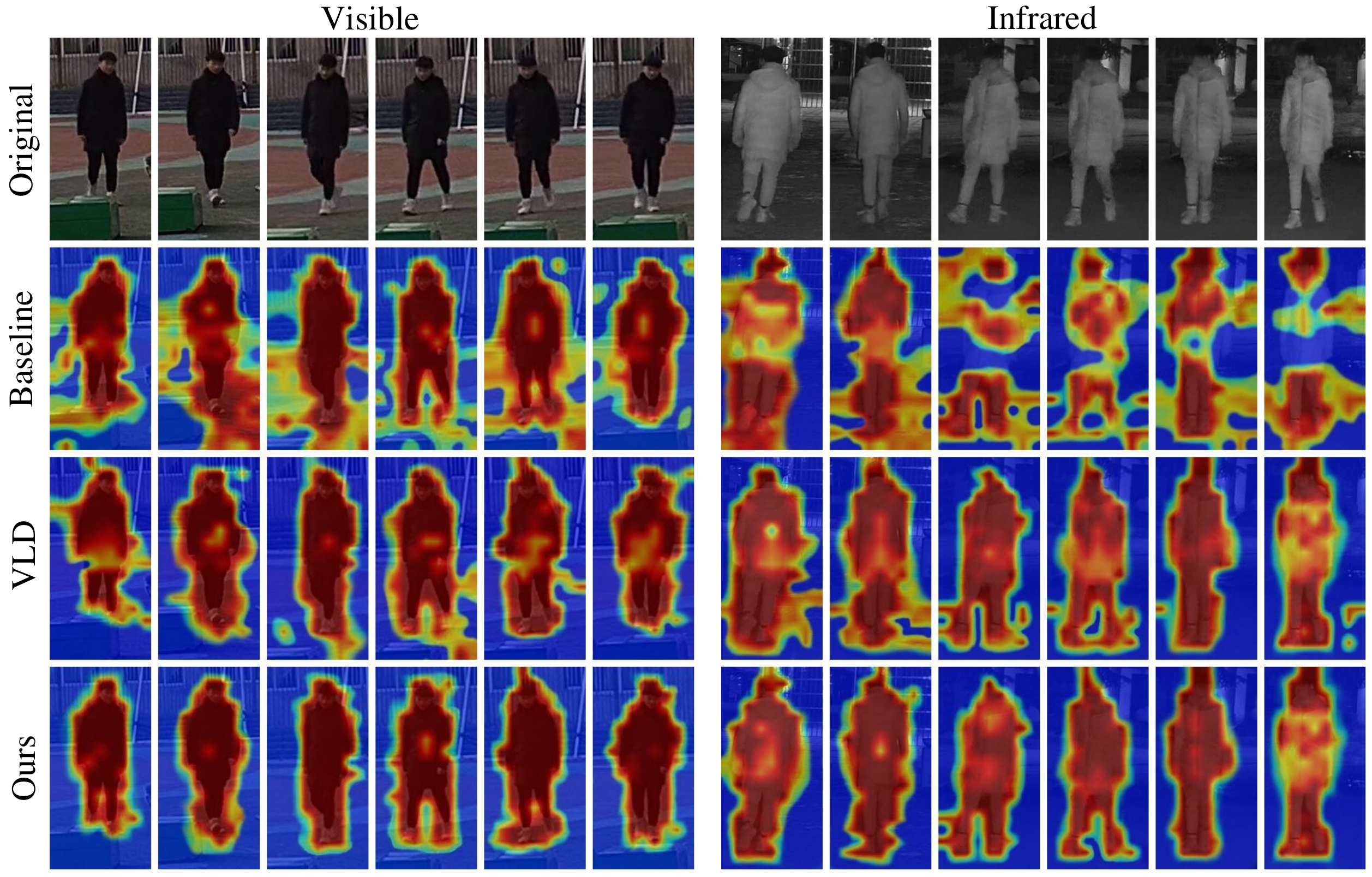} %
  \caption{CAM visualization of focus regions across different methods, warmer colors indicate stronger identity-related attention.}
  \label{cam}
\end{figure}

\begin{figure*}[!t]
\centering
  \includegraphics[width=1\textwidth]{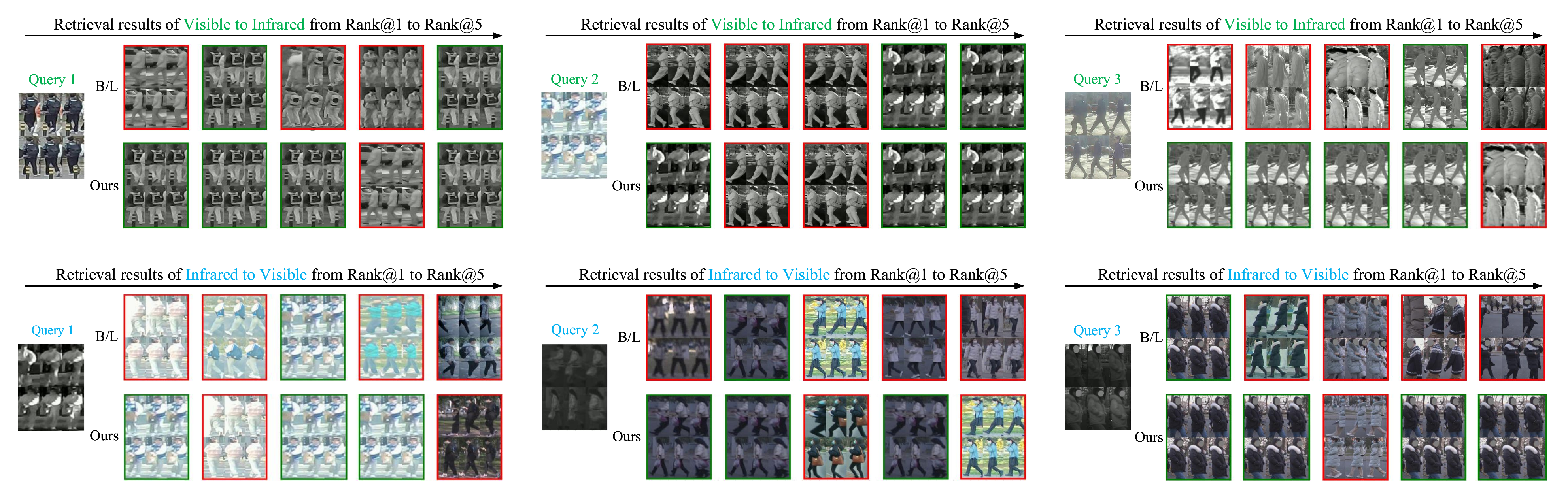} %
  \caption{Rank-5 retrieval results of some queries in V2I and I2V tasks on the BUPTCampus dataset, where B/L denote baseline method. True matches are marked with a green box, while false matches are marked with a red box.}
  \label{RANK}
\end{figure*}

\textbf{Retrieval Results Analysis}:
Fig. \ref{RANK} presents the Rank-5 retrieval results of representative queries in V2I and I2V tasks on the BUPTCampus dataset, where green/red boxes indicate true/false matches. For the V2I task, the baseline method exhibits noticeable false matches across all three queries, particularly in the Rank-1 results. In contrast, our method achieves more accurate retrieval performance, nearly all results from Rank-1 to Rank-5 are true matches. In the more challenging I2V task, our method still achieves more accurate retrieval results. For instance, the baseline only achieves a true Rank-1 match for ``query 3'', whereas our method produces true matches across all queries. Overall, these results demonstrate that our method consistently improves cross-modal retrieval accuracy and robustness under challenging conditions.

\subsection{Computational Complexity Analysis}
In this section, we analyze the computational complexity of the proposed method. As shown in Tab. \ref{Running}, our method exhibits significant advantages in terms of computational complexity, especially during the inference phase. More specifically, compared with the baseline, our method slightly increases the number of parameters and computational complexity. However, in contrast to other CLIP-based methods, our method (Inference) requires fewer parameters and achieves faster inference speed with superior performance.

\begin{table}[!t]
	\caption{Comparison of computational costs across different methods.}
	\label{Running}
	\centering
\scriptsize 
\setlength{\tabcolsep}{3.5pt}  
\begin{tabular}{l c c c c c c}
\hline
\multirow{2.5}{*}{Method}  & \multirow{2.5}{*}{Params/M}        &   \multirow{2.5}{*}{FLOPs/G} &  \multicolumn{2}{c}{I2V}  & \multicolumn{2}{c}{V2I}\\
\cmidrule(l){4-5}  \cmidrule(l){6-7}
& & & R@1 & mAP & R@1 & mAP \\
\hline
Baseline        & 86.17                  & 14.18*                 & 49.0 & 50.4 & 51.0 & 49.8 \\
TF-CLIP         & 104.26(+18.09)         & 15.16(+0.98)           & 49.4 & 51.9 & 52.5 & 51.8 \\
VLD             & 88.56(+2.39)           & 14.21(+0.03)*          & 65.3 & 63.5 & 65.8 & 63.0 \\
\hline
Ours(Training)  & 91.29(+5.12)           & 14.35(+0.17)*          & --   & --   & --   & -- \\
Ours(Inference) & \textbf{88.53(+2.36)}  & \textbf{14.19(+0.01)}* & \textbf{71.1} & \textbf{67.5} & \textbf{67.0} & \textbf{65.7} \\
\hline
\end{tabular}
\begin{tablenotes}
\footnotesize
\item  * indicates that the results were reproduced by us.
\end{tablenotes}
\end{table}

In addition to presenting experimental comparisons, we further analyze the theoretical complexity of each module under the single-sample setting. The basic vision encoder is identical to the first 8 layers of CLIP’s vision encoder, thus introducing no additional computational overhead. The STG encoder is built on the last 4 layers of CLIP’s vision encoder. Li et al. \citep{li2024zeroi2v} have proven that this operation is zero-cost and similarly introduces no additional computational overhead. The TPS encoder first performs patch shifting and then applies the self-attention mechanism. Since the shifted samples have the same number of tokens as the original samples, the additional computational overhead it introduces is: $O(N^2 D)$, here $N$ is the number of tokens, and we assume $D$ denotes the feature dimension. For the SD module, it introduces a single-query cross-modal attention, whose computational complexity is $O(ND)$. The CMI module introduces an additional self-attention mechanism, whose computational complexity is $O(N^2 D)$. From the above analysis, it can be seen that our method only introduces limited additional computational overhead while achieving significant performance improvements. This efficiency-performance balance fully demonstrates the effectiveness and practicality of the proposed method.

\subsection{Parameter Analysis}
\begin{table}[!t]
	\caption{The parameter analysis about the layers and the temporal head ratio of STG on BUPTCampus dataset.}
	\label{layer and ratio}
	\centering
\scriptsize 
\setlength{\tabcolsep}{4pt}  
		\begin{tabular}{c  c  c c c c  c c c c }
			\hline
			\multirow{2.5}{*}{Layers}&\multirow{2.5}{*}{\makecell{THR}}&\multicolumn{4}{c}{Infrared to Visible}&\multicolumn{4}{c}{Visible to Infrared}\\ 
\cmidrule(l){3-6}  \cmidrule(l){7-10}
			&&R@1&R@5&R@10&mAP&R1&R@5&R@10&mAP\\ \cline{1-10}
			0 - 11  &1/2 &39.9&58.4&64.0&36.3&35.9&54.5&62.5&33.8\\ 
            2 - 11  &1/2  &53.5&76.8&82.2&53.5&51.0&73.8&80.3&50.3\\ 
            4 - 11  &1/2  &64.2&80.8&85.8&61.0&60.9&80.5&85.6&59.8\\
            6 - 11  &1/2  &64.2&80.8&85.8&61.0&60.9&80.5&85.6&59.8\\
            \textbf{8 - 11}  &\textbf{1/2} &\textbf{71.1}&\textbf{86.0}&\textbf{89.1}&\textbf{67.5}&\textbf{67.0}&\textbf{84.4}&\textbf{89.1}&\textbf{65.7}\\
            10 - 11 &1/2  &68.2&86.0&90.2&66.0&68.4&83.6&87.9&65.1\\
            \hline
            8 - 11  &1/1  &67.2&85.6&89.5&65.3&66.6&84.0&87.9&64.1\\
            8 - 11  &1/3  &66.3&85.1&88.1&65.6&66.4&83.2&87.3&63.4\\
            8 - 11  &1/4  &67.4&84.1&88.1&65.2&66.2&82.6&86.3&63.8\\
            8 - 11  &1/5 &67.1&84.5&89.3&63.9&66.8&82.0&85.7&62.7\\
		\hline
		\end{tabular}
\end{table}
To investigate the impact of key design choices in the proposed method, we conducted parameter analysis focusing on three critical parameters: the number of STG layers integrated into the CLIP's vision encoder, the temporal head ratio (THR) value $k/h$ and the hyper-parameters $\lambda$. 

\textbf{Impact of Layers}: While keeping other components fixed, we varied the number of STG layers in the CLIP-ViT encoder. As shown in Tab. \ref{layer and ratio}, when STG is applied to all layers (0-11), the model suffers from redundant low-level feature interference, leading to suboptimal performance. As we confine STG to higher layers (e.g., (2–11), (4–11), up to (8–11)), the performance gradually improves, this aligns with our design intuition that STG is more effective for capturing discriminative spatial-temporal dependencies in high-level feature spaces. The optimal result is achieved when STG is restricted to layers (8–11): this configuration avoids low-level redundancy while fully leveraging STG's capability in high-level feature refinement. Notably, expanding STG coverage beyond this range (e.g., (10–11)) reduces performance, as it narrows the high-level feature range that STG can operate on. This analysis validates that targeted integration of STG into high-level layers (8–11) strikes the optimal balance between feature refinement and redundancy suppression.

\textbf{Impact of Temporal Head Ratio}: To explore the influence of the temporal head ratio on model performance, we fix the STG encoder to operate on layers (8–11) and adjust the ratio to 1/1, 1/2, 1/3, 1/4, and 1/5. As shown in Tab. \ref{layer and ratio}, the model achieves the best overall performance when the temporal head ratio is set to 1/2. When the ratio deviates from 1/2\,---\,either increasing to 1/1 (all heads as temporal heads) or decreasing to 1/3, 1/4, or 1/5\,---\,performance consistently declines. For instance, a ratio of 1/1 leads to a $3.9\%$ drop in I2V Rank-1 and a 0.4\% drop in V2I Rank-1, while a ratio of 1/5 results in a $3.6\%$ drop in I2V mAP and a $3.0\%$ drop in V2I mAP. This indicates that an appropriate balance between temporal and spatial heads is critical: an excessive number of temporal heads (ratio 1/1) weakens the model's ability to capture fine-grained spatial structures, while insufficient temporal heads (ratios $\leq 1/3$) limit the modeling of temporal dynamics. Thus, setting the temporal head ratio to 1/2 optimizes the trade-off between spatial structure refinement and temporal dynamic modeling, maximizing cross-modal retrieval accuracy.

\begin{figure}[!t]
\centering
\begin{subfigure}{0.32\linewidth}
  \centering
  \includegraphics[width=\linewidth,height=2cm]{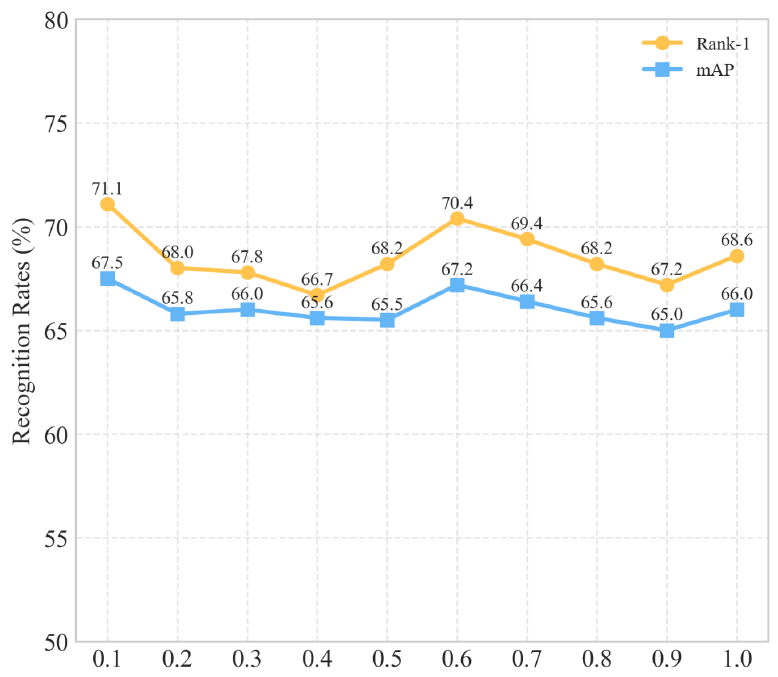}
  \caption{$\lambda_1$}
  \label{lambda_1}
\end{subfigure}
\hfill
\begin{subfigure}{0.32\linewidth}
  \centering
  \includegraphics[width=\linewidth,height=2cm]{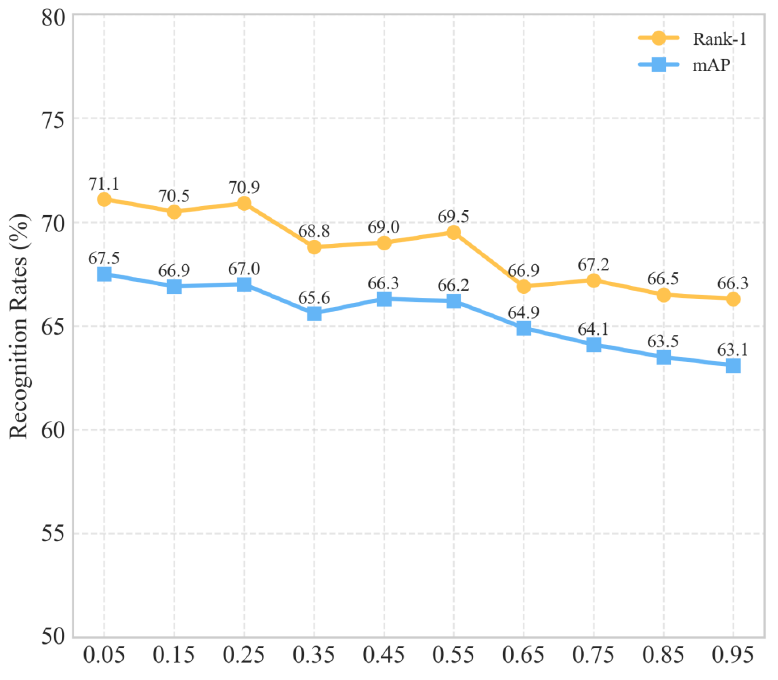}
  \caption{$\lambda_2$}
  \label{lambda_2}
\end{subfigure}
\hfill
\begin{subfigure}{0.32\linewidth}
  \centering
  \includegraphics[width=\linewidth,height=2cm]{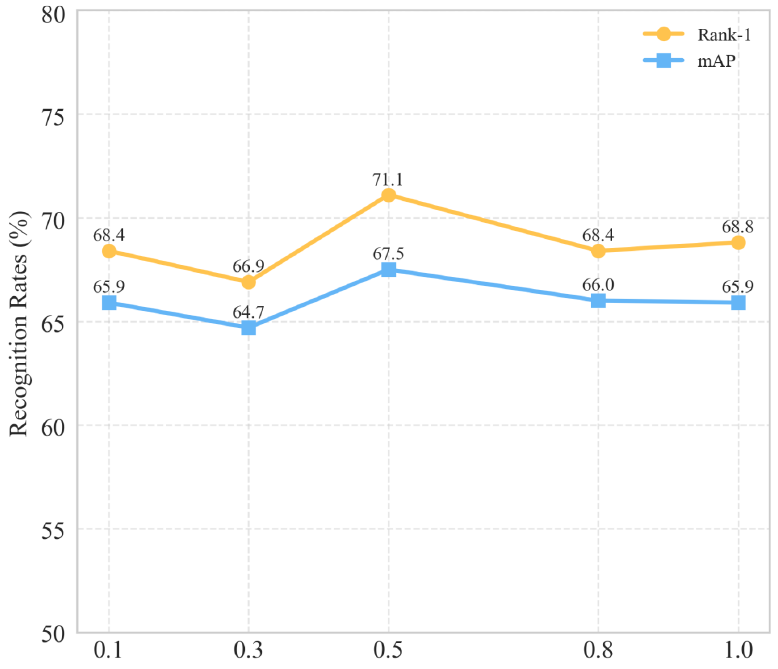}
  \caption{$\lambda_3$}
  \label{lambda_3}
\end{subfigure}
\caption{The impact of different parameters $\lambda_1$, $\lambda_2$ and $\lambda_3$ on the  BUPTCampus dataset.}
\label{lambda}
\vspace{-8pt}
\end{figure}

\textbf{Impact of $\lambda$}: 
Fig. \ref{lambda} presents the impact of hyper-parameters $\lambda_1$, $\lambda_2$, and $\lambda_3$ used in the loss function (with the other two parameters fixed during analysis). These experiments are conducted on infrared-to-visible evaluation of the BUPTCampus dataset. As shown in Fig. \ref{lambda}(a), variations in $\lambda_1$ only cause slight fluctuations in model performance while maintaining overall stability, indicating low sensitivity of the model to $\lambda_1$. In Fig. \ref{lambda}(b), the model achieves optimal performance at $\lambda_2=0.05$, and model performance shows a continuous downward trend as $\lambda_2$ increases, so the value of $\lambda_2$ needs to be constrained to avoid being excessively large. In Fig. \ref{lambda}(c), the model reaches peak performance at $\lambda_3 = 0.5$ and exhibits stable performance within the range $[0.5, 0.8]$. In summary, the recommended values for the aforementioned hyper-parameters are: $\lambda_1$ is set to $0.1$, $\lambda_2$ is set to $0.05$, and $\lambda_3$ is set to $0.5$.



\section{Conclusion} \label{sec:V}
In this paper, we propose a novel language-driven sequence-level modal-invariant representation learning method that more efficiently and effectively extends the vision-language model CLIP to the VVI-ReID field. By integrating three complementary modules: STFL, SD, and CMI, LSMRL achieves effective spatial-temporal feature extraction, coarse-to-fine cross-modal alignment, and discriminative modal-invariant representation learning while maintaining computational efficiency. Specifically, with minimal additional parameters and computational overhead, we design the STFL module based on CLIP to extract spatial-temporal features of video sequences. Then, the spatial-temporal features of both modalities are sequentially fed into the SD and CMI modules, and more discriminative and robust sequence-level modal-invariant representations are captured through sufficient cross-modal interaction. The SD module diffuses modality-shared text semantics into RGB and IR features to establish preliminary cross-modal consistency, while the CMI module leverages bidirectional cross-modal self-attention to eliminate residual modality gaps and obtain the final modal-invariant representations. Complemented by a multi-loss system that combines identity-level and modality-level losses, LSMRL enhances both the discriminability and modal invariance of sequence-level features. Extensive experiments on large-scale VVI-ReID datasets demonstrate that LSMRL outperforms SOTA methods, achieving significant improvements in Rank-1 accuracy and mAP, which indicates the great potential of pre-trained vision-language large models in enhancing ReID performance.

Despite the superior performance achieved by the proposed LSMRL method, it still has the following limitations. First, LSMRL is designed based on the coarse-grained backbone CLIP, making it struggle to effectively align local semantics. Second, we infer that the modal-invariant features obtained by LSMRL still contain modality-specific information, which impairs the model's discriminative ability and causes interference in VVI-ReID. Therefore, in future work, we will introduce fine-grained alignment techniques and design feature decoupling networks to further enhance the method's discriminative ability and ReID performance.


\section*{Acknowledgments}
This document is the results of the research project funded by the Shandong Provincial Natural Science Foundation (Nos. ZR2025MS1004, ZR2025QC647) and the National Natural Science Foundation of China (Nos. 62471202, 62406143, 62101213).



\bibliographystyle{IEEEtran}
\bibliography{reference}

@inproceedings{lin2022learning,
  title={Learning modal-invariant and temporal-memory for video-based visible-infrared person re-identification},
  author={Lin, Xinyu and Li, Jinxing and Ma, Zeyu and Li, Huafeng and Li, Shuang and Xu, Kaixiong and Lu, Guangming and Zhang, David},
  booktitle={Proceedings of the IEEE/CVF Conference on Computer Vision and Pattern Recognition},
  pages={20973--20982},
  year={2022}
}

@article{du2023video,
  title={Video-based visible-infrared person re-identification with auxiliary samples},
  author={Du, Yunhao and Lei, Cheng and Zhao, Zhicheng and Dong, Yuan and Su, Fei},
  journal={IEEE Transactions on Information Forensics and Security},
  volume={19},
  pages={1313--1325},
  year={2023},
  publisher={IEEE}
}

@article{ye2021deep,
  title={Deep learning for person re-identification: A survey and outlook},
  author={Ye, Mang and Shen, Jianbing and Lin, Gaojie and Xiang, Tao and Shao, Ling and Hoi, Steven CH},
  journal={IEEE Transactions on Pattern Analysis and Machine Intelligence},
  volume={44},
  number={6},
  pages={2872--2893},
  year={2021},
  publisher={IEEE}
}

@ARTICLE{feng2024cross,
  author={Feng, Yujian and Chen, Feng and Yu, Jian and Ji, Yimu and Wu, Fei and Liu, Tianliang and Liu, Shangdong and Jing, Xiao-Yuan and Luo, Jiebo},
  journal={IEEE Transactions on Multimedia}, 
  title={Cross-Modality Spatial-Temporal Transformer for Video-Based Visible-Infrared Person Re-Identification}, 
  year={2024},
  volume={26},
  number={},
  pages={6582-6594}
}

@inproceedings{li2023clip,
  title={Clip-reid: exploiting vision-language model for image re-identification without concrete text labels},
  author={Li, Siyuan and Sun, Li and Li, Qingli},
  booktitle={Proceedings of the AAAI Conference on Artificial Intelligence},
  volume={37},
  number={1},
  pages={1405--1413},
  year={2023}
}

@inproceedings{li2024zeroi2v,
  title={Zeroi2v: Zero-cost adaptation of pre-trained transformers from image to video},
  author={Li, Xinhao and Zhu, Yuhan and Wang, Limin},
  booktitle={European Conference on Computer Vision},
  pages={425--443},
  year={2024},
  organization={Springer}
}

@inproceedings{ye2020dynamic,
	title={Dynamic dual-attentive aggregation learning for visible-infrared person re-identification},
	author={Ye, Mang and Shen, Jianbing and J. Crandall, David and Shao, Ling and Luo, Jiebo},
	booktitle={Computer Vision--ECCV 2020: 16th European Conference, Glasgow, UK, August 23--28, 2020, Proceedings, Part XVII 16},
	pages={229--247},
	year={2020},
	organization={Springer}
}

@inproceedings{park2021learning,
	title={Learning by aligning: Visible-infrared person re-identification using cross-modal correspondences},
	author={Park, Hyunjong and Lee, Sanghoon and Lee, Junghyup and Ham, Bumsub},
	booktitle={Proceedings of the IEEE/CVF International Conference on Computer Vision},
	pages={12046--12055},
	year={2021}
}

@inproceedings{ye2021channel,
	title={Channel augmented joint learning for visible-infrared recognition},
	author={Ye, Mang and Ruan, Weijian and Du, Bo and Shou, Mike Zheng},
	booktitle={Proceedings of the IEEE/CVF International Conference on Computer Vision},
	pages={13567--13576},
	year={2021}
}

@inproceedings{tian2021farewell,
	title={Farewell to mutual information: Variational distillation for cross-modal person re-identification},
	author={Tian, Xudong and Zhang, Zhizhong and Lin, Shaohui and Qu, Yanyun and Xie, Yuan and Ma, Lizhuang},
	booktitle={Proceedings of the IEEE/CVF Conference on Computer Vision and Pattern Recognition},
	pages={1522--1531},
	year={2021}
}

@inproceedings{zhang2023diverse,
	title={Diverse embedding expansion network and low-light cross-modality benchmark for visible-infrared person re-identification},
	author={Zhang, Yukang and Wang, Hanzi},
	booktitle={Proceedings of the IEEE/CVF Conference on Computer Vision and Pattern Recognition},
	pages={2153--2162},
	year={2023}
}

@inproceedings{wu2021discover,
	title={Discover cross-modality nuances for visible-infrared person re-identification},
	author={Wu, Qiong and Dai, Pingyang and Chen, Jie and Lin, Chia-Wen and Wu, Yongjian and Huang, Feiyue and Zhong, Bineng and Ji, Rongrong},
	booktitle={Proceedings of the IEEE/CVF Conference on Computer Vision and Pattern Recognition},
	pages={4330--4339},
	year={2021}
}

@inproceedings{qiu2024high,
  title={High-order structure based middle-feature learning for visible-infrared person re-identification},
  author={Qiu, Liuxiang and Chen, Si and Yan, Yan and Xue, Jing-Hao and Wang, Da-Han and Zhu, Shunzhi},
  booktitle={Proceedings of the AAAI Conference on Artificial Intelligence},
  volume={38},
  number={5},
  pages={4596--4604},
  year={2024}
}

@article{yuan2024unbiased,
  title={Unbiased feature learning with causal intervention for visible-infrared person re-identification},
  author={Yuan, Bowen and Lu, Jiahao and You, Sisi and Bao, Bing-kun},
  journal={ACM Transactions on Multimedia Computing, Communications and Applications},
  volume={20},
  number={10},
  pages={1--20},
  year={2024},
  publisher={ACM New York, NY}
}

@article{zuo2025spatio,
  title={Spatio-temporal information mining and fusion feature-guided modal alignment for video-based visible-infrared person re-identification},
  author={Zuo, Zhigang and Li, Huafeng and Zhang, Yafei and Xie, Minghong},
  journal={Image and Vision Computing},
  volume={157},
  pages={105518},
  year={2025},
  publisher={Elsevier}
}

@inproceedings{zhou2023video,
  title={Video-based visible-infrared person re-identification via style disturbance defense and dual interaction},
  author={Zhou, Chuhao and Li, Jinxing and Li, Huafeng and Lu, Guangming and Xu, Yong and Zhang, Min},
  booktitle={Proceedings of the 31st ACM International Conference on Multimedia},
  pages={46--55},
  year={2023}
}

@inproceedings{yu2024tf,
  title={Tf-clip: Learning text-free clip for video-based person re-identification},
  author={Yu, Chenyang and Liu, Xuehu and Wang, Yingquan and Zhang, Pingping and Lu, Huchuan},
  booktitle={Proceedings of the AAAI Conference on Artificial Intelligence},
  volume={38},
  number={7},
  pages={6764--6772},
  year={2024}
}

@inproceedings{zhang2021towards,
	title={Towards a unified middle modality learning for visible-infrared person re-identification},
	author={Zhang, Yukang and Yan, Yan and Lu, Yang and Wang, Hanzi},
	booktitle={Proceedings of the 29th ACM international conference on Multimedia},
	pages={788--796},
	year={2021}
}

@inproceedings{yang2022learning,
	title={Learning with twin noisy labels for visible-infrared person re-identification},
	author={Yang, Mouxing and Huang, Zhenyu and Hu, Peng and Li, Taihao and Lv, Jiancheng and Peng, Xi},
	booktitle={Proceedings of the IEEE/CVF Conference on Computer Vision and Pattern Recognition},
	pages={14308--14317},
	year={2022}
}

@ARTICLE{li2025video,
  author={Li, Shuang and Leng, Jiaxu and Kuang, Changjiang and Tan, Mingpi and Gao, Xinbo},
  journal={IEEE Transactions on Information Forensics and Security}, 
  title={Video-Level Language-Driven Video-Based Visible-Infrared Person Re-Identification}, 
  year={2025},
  volume={20},
  number={},
  pages={5505-5520}
}

@InProceedings{xiang2022TPS,
author="Xiang, Wangmeng
and Li, Chao
and Wang, Biao
and Wei, Xihan
and Hua, Xian-Sheng
and Zhang, Lei",
title="Spatiotemporal Self-attention Modeling with Temporal Patch Shift for Action Recognition",
booktitle="Computer Vision -- ECCV 2022",
year="2022",
publisher="Springer Nature Switzerland",
address="Cham",
pages="627--644",
isbn="978-3-031-20062-5"
}

@ARTICLE{Feng2023MDloss,
  author={Feng, Yujian and Yu, Jian and Chen, Feng and Ji, Yimu and Wu, Fei and Liu, Shangdon and Jing, Xiao-Yuan},
  journal={IEEE Transactions on Multimedia}, 
  title={Visible-Infrared Person Re-Identification via Cross-Modality Interaction Transformer}, 
  year={2023},
  volume={25},
  number={},
  pages={7647-7659}
}

@InProceedings{Lu2023MSELloss,
author="Lu, Hu and Zou, Xuezhang and Zhang, Pingping",
title="Learning Progressive Modality-Shared Transformers for Effective Visible-Infrared Person Re-identification",
booktitle="Proceedings of the AAAI Conference on Artificial Intelligence",
year="2023",
volume = {37},
nuber = {2},
pages="1835-1843"
}

@article{Wang2025DIRL,
author = {Wang, Jiahe and Gao, Xizhan and Niu, Sijie and Zhao, Hui and Feng, Guang},
title = {DIRL: Learning Discriminative ID-Related Representations for Video Visible-Infrared Person ReID},
year = {2025},
issue_date = {August 2025},
publisher = {Association for Computing Machinery},
address = {New York, NY, USA},
volume = {21},
number = {8},
issn = {1551-6857},
journal = {ACM Transactions on Multimedia Computing, Communications and Applications},
month = aug,
articleno = {238},
pages = {238:1-16}
}

@article{Wang2025TAE,
title = {Learning discriminative features via deep metric learning for video-based person re-identification},
journal = {Expert Systems with Applications},
volume = {286},
pages = {128123},
year = {2025},
issn = {0957-4174},
author = {Jiahe Wang and Xizhan Gao and Sijie Niu and Hui Zhao and Guang Feng and Jiaxin Lin}
}

@inproceedings{zheng2016mars,
  title={Mars: A video benchmark for large-scale person re-identification},
  author={Zheng, Liang and Bie, Zhi and Sun, Yifan and Wang, Jingdong and Su, Chi and Wang, Shengjin and Tian, Qi},
  booktitle={European Conference on Computer Vision},
  pages={868--884},
  year={2016},
  organization={Springer}
}

@article{MA2024128479,
title = {A review on video person re-identification based on deep learning},
journal = {Neurocomputing},
volume = {609},
pages = {128479},
year = {2024},
issn = {0925-2312},
author = {Haifei Ma and Canlong Zhang and Yifeng Zhang and Zhixin Li and Zhiwen Wang and Chunrong Wei}
}

@ARTICLE{Li2023IBAN,
  author={Li, Huafeng and Liu, Minghui and Hu, Zhanxuan and Nie, Feiping and Yu, Zhengtao},
  journal={IEEE Transactions on Circuits and Systems for Video Technology}, 
  title={Intermediary-Guided Bidirectional Spatial–Temporal Aggregation Network for Video-Based Visible-Infrared Person Re-Identification}, 
  year={2023},
  volume={33},
  number={9},
  pages={4962-4972}
}

@inproceedings{radford2021learning,
  title={Learning transferable visual models from natural language supervision},
  author={Radford, Alec and Kim, Jong Wook and Hallacy, Chris and Ramesh, Aditya and Goh, Gabriel and Agarwal, Sandhini and Sastry, Girish and Askell, Amanda and Mishkin, Pamela and Clark, Jack and others},
  booktitle={International Conference on Machine Learning},
  pages={8748--8763},
  year={2021},
  organization={PmLR}
}

@ARTICLE{Wu2024BOV,
  author={Wu, Zuxuan and Weng, Zejia and Peng, Wujian and Yang, Xitong and Li, Ang and Davis, Larry S. and Jiang, Yu-Gang},
  journal={IEEE Transactions on Pattern Analysis and Machine Intelligence}, 
  title={Building an Open-Vocabulary Video CLIP Model With Better Architectures, Optimization and Data}, 
  year={2024},
  volume={46},
  number={7},
  pages={4747-4762},
  doi={10.1109/TPAMI.2024.3357503}
}

@ARTICLE{yang2025clip4vireid,
  author={Xiaomei Yang and Xizhan Gao and Sijie Niu and Fa Zhu and Guang Feng and Xiaofeng Qu and David Camacho},
  journal={arXiv}, 
  title={CLIP4VI-ReID: Learning Modality-shared Representations via CLIP Semantic Bridge for Visible-Infrared Person Re-identification}, 
  year={2025},
  volume={},
  number={},
  pages={2511.10309},
  url={https://arxiv.org/abs/2511.10309}
}

@ARTICLE{Hu2025CLIPMC,
  author={Hu, Gang and Lv, Yafei and Zhang, Jianting and Wu, Qian and Wen, Zaidao},
  journal={IEEE Transactions on Multimedia}, 
  title={CLIP-Based Modality Compensation for Visible-Infrared Image Re-Identification}, 
  year={2025},
  volume={27},
  number={},
  pages={2112-2126},
  doi={10.1109/TMM.2024.3521764}}

@article{Yu2024CSDN,
  title={CLIP-Driven Semantic Discovery Network for Visible-Infrared Person Re-Identification},
  author={Xiaoyan Yu and Neng Dong and Liehuang Zhu and Hao Peng and Dapeng Tao},
  journal={IEEE Transactions on Multimedia},
  year={2024},
  volume={27},
  pages={4137-4150}
}

@inproceedings{Chen2023CCLNet,
author = {Chen, Zhong and Zhang, Zhizhong and Tan, Xin and Qu, Yanyun and Xie, Yuan},
title = {Unveiling the Power of CLIP in Unsupervised Visible-Infrared Person Re-Identification},
year = {2023},
isbn = {9798400701085},
publisher = {Association for Computing Machinery},
address = {New York, NY, USA},
booktitle = {Proceedings of the 31st ACM International Conference on Multimedia},
pages = {3667–3675},
numpages = {9},
location = {Ottawa ON, Canada},
series = {MM '23}
}

@ARTICLE{Wu2024OpenVCLIP,
  author={Wu, Zuxuan and Weng, Zejia and Peng, Wujian and Yang, Xitong and Li, Ang and Davis, Larry S. and Jiang, Yu-Gang},
  journal={IEEE Transactions on Pattern Analysis and Machine Intelligence}, 
  title={Building an Open-Vocabulary Video CLIP Model With Better Architectures, Optimization and Data}, 
  year={2024},
  volume={46},
  number={7},
  pages={4747-4762},
  doi={10.1109/TPAMI.2024.3357503}}

@article{Yu2025CLIMB-ReID, 
title={CLIMB-ReID: A Hybrid CLIP-Mamba Framework for Person Re-Identification}, 
volume={39}, 
number={9},
journal={Proceedings of the AAAI Conference on Artificial Intelligence}, 
author={Yu, Chenyang and Liu, Xuehu and Zhu, Jiawen and Wang, Yuhao and Zhang, Pingping and Lu, Huchuan}, 
year={2025}, 
month={Apr.}, 
pages={9589-9597} 
}

@ARTICLE{Li2025UDG,
  author={Li, Huafeng and Liu, Yaoxin and Zhang, Yafei and Li, Jinxing and Yu, Zhengtao},
  journal={IEEE Transactions on Information Forensics and Security}, 
  title={Breaking the Paired Sample Barrier in Person Re-Identification: Leveraging Unpaired Samples for Domain Generalization}, 
  year={2025},
  volume={20},
  number={},
  pages={2357-2371},
  doi={10.1109/TIFS.2025.3543040}
}

@article{Wang2025SVLLReID,
title = {Image re-identification: Where self-supervision meets vision-language learning},
journal = {Image and Vision Computing},
volume = {154},
pages = {105415},
year = {2025},
issn = {0262-8856},
author = {Bin Wang and Yuying Liang and Lei Cai and Huakun Huang and Huanqiang Zeng}
}

@article{Zhou2025HDGI,
title = {Hierarchical disturbance and Group Inference for video-based visible-infrared person re-identification},
journal = {Information Fusion},
volume = {117},
pages = {102882},
year = {2025},
issn = {1566-2535},
author = {Chuhao Zhou and Yuzhe Zhou and Tingting Ren and Huafeng Li and Jinxing Li and Guangming Lu}
}

@misc{yu2025xreid,
      title={X-ReID: Multi-granularity Information Interaction for Video-Based Visible-Infrared Person Re-Identification}, 
      author={Chenyang Yu and Xuehu Liu and Pingping Zhang and Huchuan Lu},
      year={2025},
      eprint={2511.17964},
      archivePrefix={arXiv},
      primaryClass={cs.CV},
      url={https://arxiv.org/abs/2511.17964}, 
}
 
%

\newpage

\section{Biography Section}
If you have an EPS/PDF photo (graphicx package needed), extra braces are
 needed around the contents of the optional argument to biography to prevent
 the LaTeX parser from getting confused when it sees the complicated
 $\backslash${\tt{includegraphics}} command within an optional argument. (You can create
 your own custom macro containing the $\backslash${\tt{includegraphics}} command to make things
 simpler here.)
 
\vspace{11pt}




\vfill

\end{document}